\documentclass[final]{nesy2026} % Uncomment to include author names

\usepackage{cleveref}
\usepackage{multirow}
\usepackage{longtable}% for long tables
\usepackage{float}
\usepackage{titlesec}
\usepackage{float}
\titlespacing*{\section}{0pt}{4pt}{3pt}
\titlespacing*{\subsection}{0pt}{4pt}{2pt}
\titlespacing*{\subsubsection}{0pt}{3pt}{1pt}
\usepackage{booktabs}
\usepackage{microtype}
\usepackage{siunitx}
\usepackage[skip=2pt]{caption}
\theorembodyfont{\upshape}
\theoremheaderfont{\scshape}
\theorempostheader{:}
\theoremsep{\newline}

\usepackage{xcolor}

\title[Do CNNs Internally Represent Real and Fake Images Differently?]{Do CNNs Internally Represent Real and Fake Images Differently? A Hidden-Layer Analysis}

 \author{\Name{Moumita Sen Sarma} \Email{moumita@ksu.edu}\\
 \Name{Pascal Hitzler} \Email{hitzler@ksu.edu}\\
 \Name{Eugene Y. Vasserman} \Email{eyv@ksu.edu}\\
 \addr Department of Computer Science, Kansas State University, Manhattan, Kansas, USA}
\begin{document}

\maketitle
\vspace{-1em}
\begin{abstract}
%\mss{modified abstract reflecting the new analysis}

Fake/synthetic images are increasingly prevalent, but it remains unclear whether Convolutional Neural Networks (CNNs) process real and fake images in the same internal manner. This work examines the hypothesis that CNNs represent real and fake images differently, such that fake images induce different hidden-layer activation patterns even when semantic content is preserved. The hypothesis is evaluated in scene recognition settings using trained CNN models. Dense-layer activations are extracted, and neurosymbolic methods assign semantic labels to selected neurons. For each real test image, corresponding fake images are generated with similar semantic content using object-label-guided text-to-image and image-to-image generation based on Stable Diffusion variants. Paired real-fake activation patterns are then compared statistically. Additional experiments with another dataset, CNN architecture, generative model, and JPEG/blur degradation analysis assess robustness. Results suggest that fake images evoke different hidden-neuron activations, and these differences are not explained only by simple image degradation.
 Overall, the findings indicate that real and fake images differ in CNN hidden-layer activation behavior at least in some settings, which opens the door for follow-up work on making use of this different behavior to improve fake image detection.

\end{abstract}

\section{Introduction}
\label{sec:intro}
%\vspace{-.5em}
Fake images generated by diffusion and other models are increasingly used in vision systems for data augmentation, rare scenario simulation, and reduced annotation cost~\citep{doi:10.1142/S0129065725500522, 11079597, jimaging8110310}. At the same time, their widespread presence in media and automated systems raises serious concerns. Moreover, generative models may introduce subtle structural artifacts or distributional inconsistencies that are difficult to detect~\citep{NEURIPS2023_505df5ea}. As fake images become visually indistinguishable from real ones~\citep{doi:10.1073/pnas.2120481119}, the boundary between original and generated data blurs, posing risks to reliability, forensic integrity, and trust in AI systems~\citep{9115874, app131910980, article}.
Despite these concerns, fake and real images are often treated equivalently in deep learning pipelines. It remains unclear whether
%\mss{changed CNNs from DNNs (it was overstated)}
CNNs internally process fake images in the same way as real ones. Although CNNs achieve strong performance in visual recognition~\citep{Zhao2024ARO}, their hidden-layer representations remain opaque~\citep{HAAR2023105606}. Existing XAI methods primarily focus on input-output behavior and saliency maps~\citep{8237336, zheng2022shapcam}, without examining how semantic information is encoded in hidden neurons or whether internal representations differ between real and fake images.

%\eyv{@Moumita, avoid using citations as nouns. For instance, instead of ``In~\citep{G2024DetectingAI,HasanAbir2023DetectingDI}, CNN models are trained to classify real and fake face images on a large-scale Kaggle dataset ($\sim$140K images), using Grad-CAM and LIME to visualize regions influencing predictions.'' consider ``Prior work has used CNNs to classify real and fake face images, trained on a large-scale Kaggle dataset ($\sim$140K images), and using Grad-CAM and LIME to visualize regions influencing predictions~\citep{G2024DetectingAI,HasanAbir2023DetectingDI}.''} 
%\mss{Resolved}

%\pascal{I actually disagree with Eugene. I know what he says is "correct" typesetting, but I think it leads to convoluted reading, so I often deliberately ignore what's considered "correct". In the end, I personally don't care :) }
%\mss{I tried to follow the correct typesetting as much as possible.}

%\eyv{OK, I do see your point, especially for this specific paper format, as (subjectively) author/year citations feel better suited for use as nouns than numbered citations.}

Several prior works have investigated fake image and deepfake detection using CNN-based classifiers combined with post-hoc explainability methods. For example, CNN detectors have been evaluated on large-scale real/fake face datasets, CIFAKE, and FaceForensics++-based deepfake datasets, with Grad-CAM~\citep{8237336}, LIME~\citep{ribeiro2016should}, LRP~\citep{Bach2015OnPE}, and SHAP~\citep{NIPS2017_7062} used to highlight image regions that influence model predictions~\citep{G2024DetectingAI,HasanAbir2023DetectingDI,Bird2023CIFAKEIC,Malolan2020ExplainableDD,11316647}. These studies are useful for visualizing where a detector attends, but their explanations are mainly output-level and pixel- or region-based. They do not directly examine whether real and fake images produce systematically different hidden-layer activation patterns or neuron-level representations.
Other works have focused on evaluating the reliability of visual explanations. For example, perturbation-based evaluation has been used to test whether highlighted regions affect detector performance~\citep{10.1145/3643491.3660292}. While this provides a more quantitative assessment of saliency quality, the interpretation remains tied to perturbation choices and does not provide semantic analysis of internal neurons. More recently, vision-language models have been evaluated for fake image reasoning using FakeBench~\citep{Li2024FakeBenchPE}; however, such explanations are language-based and model-dependent, focusing on generated reasoning rather than internal visual representations.

Collectively, prior work emphasizes detection accuracy and output-level explanations. However, it remains largely unexplored how hidden-layer activation patterns differ between real and fake images when the semantic content is kept the same.

%\pascal{We cannot use "our" because it's double-blind. Please check the whole paper to make sure that we don't lift anonymity anywhere. E.g. also check how we share data, it needs to be via an anonymous URL and site.} \eyv{It's slightly more difficult than simply replacing ``our'' and ``we''. Now that we can't point out that the prior work is ours, we ened to be even more careful in showing novelty/difference from prior work.} 

%\mss{Removed "we", "our". Shared data is published in an anonymous url, given in footnote in pg-3}

This work extends the concept induction-based framework for neuron-level interpretability introduced in~\citep{10.1007/978-3-031-71170-1_12}, where hidden neurons are semantically grounded using ECII~\citep{Sarker_Hitzler_2019} and statistically validated on ADE20K
% \pascal{add citation} 
~\citep{10.1007/978-3-031-71170-1_12}. The framework was later transferred to SUN2012 
% \pascal{add citation}
dataset~\citep{sarma2026casestudyconceptinduction}, demonstrating its generalizability across benchmarks. Building on this foundation, it is examined whether neurons tagged with specific concepts for real images exhibit similar activation behavior for semantically matched fake counterparts.\footnote{The source code, along with input datasets and generated outputs are available online at: \url{https://github.com/Moumita-Sen-Sarma/Hidden-neuron-analysis-on-fake-image}} The central hypothesis is that CNNs internally represent real and fake images differently, with fake images inducing distinct hidden-layer activation patterns. Paired real-fake analysis reveals significant representation-level differences in neuron activations.

%\pascal{Should also mention your K-Cap poster on transfer to Sun2012.} \mss{Resolved}

%\vspace{-1em}
\section{Background}
%\pascal{You can probably remove most or all of this and instead refer to the Dalal 2024 and the Sen Sarma 2012 K-Cap papers. Perhaps summarize the most needed in a few sentences, as part of a later section.} \mss{Resolved}
%\vspace{-.5em}

\noindent \textbf{Concept Induction-Based Neuron Interpretability Framework:} 
This work builds upon the concept induction-based neuron interpretability framework introduced by~\citep{10.1007/978-3-031-71170-1_12} and subsequently transferred to the SUN2012 dataset in
% \pascal{add citation} 
~\citep{sarma2026casestudyconceptinduction}. In the prior framework, annotated objects are mapped to lexical matches in the Wikipedia Knowledge Graph~\citep{10.1007/978-3-030-65384-2_6} to form a background ontology, over which concept induction induces logical class expressions that separate positive and negative activation sets. The induced neuron labels were validated through web-sourced image confirmation and statistical testing, demonstrating that robust neuron-concept associations generalize across benchmarks. 
% \pascal{Later (Section 3.2) you refer to "confirmed neurons" -- that notion isn't explained yet. I think you could explain it here?} 
In this regard, a neuron's label is confirmed when Target Level Activation (TLA), defined as the percentage of retrieved concept-related images activating the neuron above threshold, is at least 80\%, and a Mann-Whitney U test (p $<$ 0.05, negative z-score) indicates significantly stronger activations for target images.
For our work, the complete pipeline is re-executed to ensure consistency with the experimental setting. Although slight differences in the set of confirmed neurons are observed, attributable to model retraining, the overall interpretability outcomes remained closely aligned with the results of
% \pascal{add citation}
~\citep{sarma2026casestudyconceptinduction}.

\noindent \textbf{Image Generation Model:} 
% \mss{As I have done additional experiment with FLUX.1 image generation model, should i add its details here? I am concerned about the page limit.} \pascal{Since you've already done the work, I'd suggest to at least mention it in a sentence, something like ``We also conducted the same/similar experiments with Flux.1 [citation] with similar results / somewhat different results detailed in Appendix xyz'' or whatever wording is adequate.} \mss{Done}
\noindent Image generation models synthesize realistic images from learned data distributions and are typically conditioned on text or images. They are widely used for content creation, data augmentation, and robustness evaluation~\citep{9878449, podell2023sdxlimprovinglatentdiffusion}. Stable Diffusion~\citep{9878449} is a latent diffusion model that generates images by iteratively denoising a compressed latent representation guided by text embeddings. Compared to GAN-based and proprietary large-scale models, it offers stable training, strong diversity, competitive image quality, and computational efficiency while remaining open-source and reproducible~\citep{10625113, podell2023sdxlimprovinglatentdiffusion}. Its support for both text-conditioned and image-conditioned generation makes it suitable for structured comparative analysis. 
Architecturally, Stable Diffusion consists of a Variational Autoencoder (VAE), a U-Net denoising network, and a text encoder. The VAE encodes images into a latent space and decodes them back to pixel space, while the U-Net iteratively removes noise in the latent domain conditioned on text embeddings. In text-based generation, the process begins from random latent noise. This latent diffusion design reduces computational cost while maintaining high visual fidelity and semantic consistency.
Similar experiments are also conducted with FLUX.1~\citep{flux2024}, a recent open-weight rectified-flow-based image generation model, and comparable results are reported in Appendix~\ref{apd:second}. It is a 12B-parameter rectified flow transformer and is trained using guidance distillation, which is intended to improve inference efficiency while maintaining strong image quality and prompt-following ability.
\begin{figure}[!ht]
    \centering
    \includegraphics[width=0.7\textwidth]{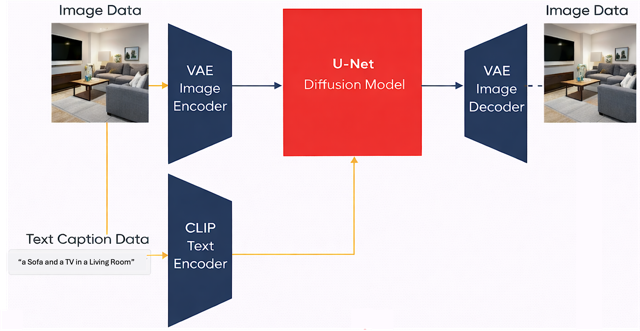}
    \caption{Stable Diffusion Model Architecture. Adapted from~\cite{ritika_diffusion}}
    \label{fig:sd}
\end{figure}

\begin{figure}[thb]
    \centering
    \includegraphics[width=\textwidth]{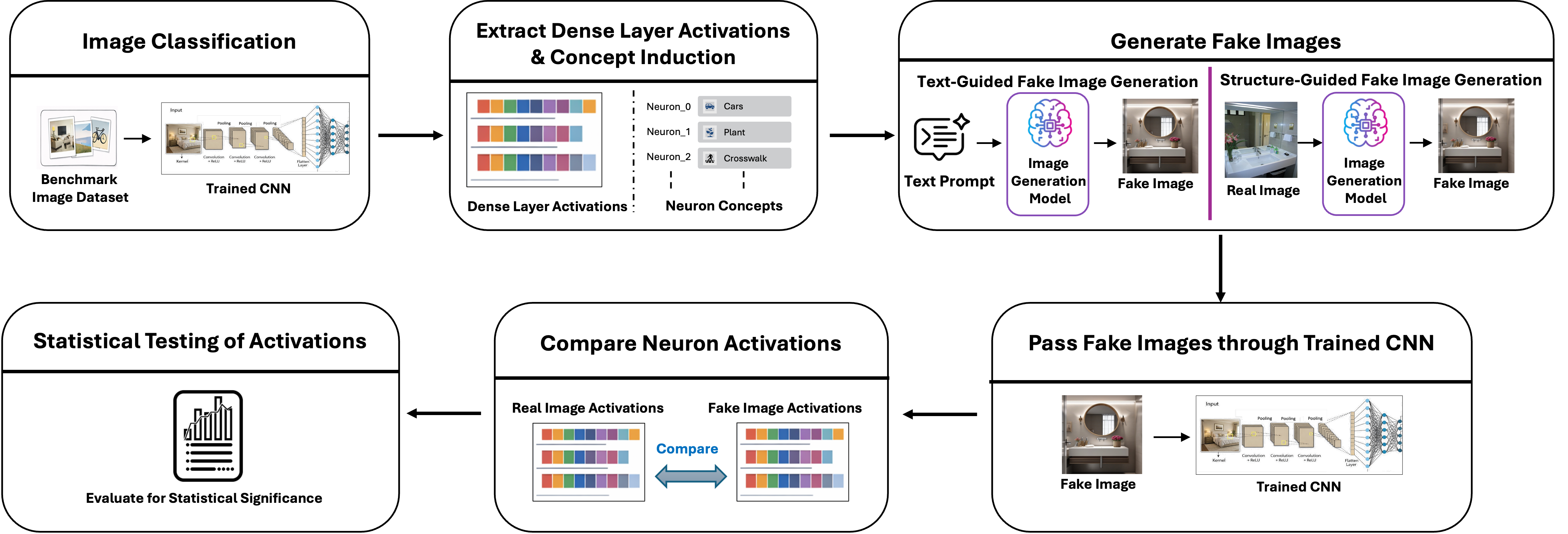}
    \caption{
    % \mss{modified the `Generate Fake images' part of the figure as per comment of reviewer 4.} 
    Workflow for comparing real and generated images through hidden-layer neuron activations. In this approach, fake images are generated using both text-only and image along with text settings before activation-level statistical comparison.}
    \label{fig:methodology}
\end{figure}

%\vspace{-1em}
\section{Methodology}
%\vspace{-.5em}
%\pascal{You can probably remove a lot of the content in the Methodology section by refering to your K-Cap 2012 poster paper, or condense it to bare bones, and refering the reader to your K-Cap and the Dalal paper. You can also save a \emph{lot} of vertical space by getting rid of sub-headings.}

%\pascal{Alternatively, you could write it all up in very detail (as you do here), and put this online somewhere as a pdf (anonymous!) then refer to it in your briefer paper, saying "refer to the online version for further details". Of course the submitted version still needs to be self-contained.}
%\mss{Resolved}

\noindent In this work, a six-step framework is employed to analyze activation differences between real and fake images (depicted in Figure~\ref{fig:methodology}). The initial stages, including CNN training, dense-layer activation extraction, and concept induction for assigning semantic labels to neurons, are conducted following the approach established in~\citep{10.1007/978-3-031-71170-1_12, sarma2026casestudyconceptinduction}. Building on this foundation, fake images are generated for each real image and passed through the same trained CNN to obtain corresponding activations. Paired real-fake activation patterns are then compared using statistical analysis to identify significant representation-level differences. 

\subsection{Generate Fake Images}
%\vspace{-.5em}
To analyze hidden-layer activation differences with matched semantic content, fake images are generated using two approaches. In the text-guided approach, fake images are generated from object annotation-based prompt, while in the structure-guided approach, the real image is provided along with the prompt to produce a structurally informed fake image.

\begin{figure}[!htb]
    \centering
    \includegraphics[height=11cm,width=15cm,trim={0 4mm 0 4mm},clip]{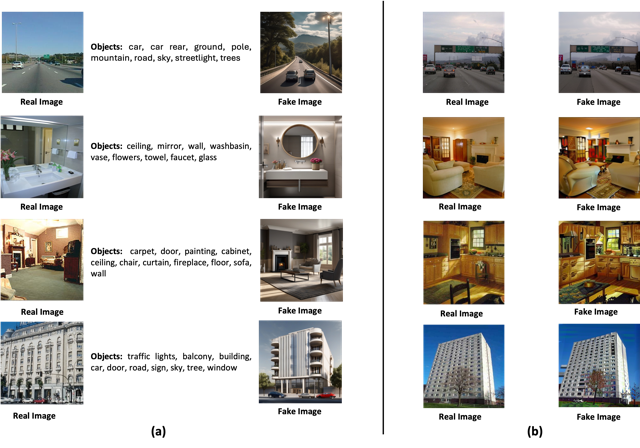}
    \caption{Sample fake output images using SUN2012 and Stable Diffusion variants; (a): from Text-Guided Fake Image Generation using ground-truth object annotations; (b): from Structure-Guided Fake Image Generation. Additional examples are depicted in Figure~\ref{fig:fake_sample_sun_stab} in Appendix~\ref{apd:samples}.}
    \label{fig:output_fakes}
\end{figure}

\vspace{-.8em}
%\noindent \textbf{Text-Guided Fake Image Generation}
\paragraph{Text-Guided Fake Image Generation}
In this approach,  Stable Diffusion XL Base 1.0 is employed, a  recent and higher-capacity version with improved text understanding and image fidelity, in text-to-image mode \citep{podell2023sdxlimprovinglatentdiffusion}. 
Fake images are generated using a structured prompt of the form:
\textit{``A realistic photo of a $ \langle \text{scene\_class} \rangle$ interior, natural lighting, high detail, sharp focus, containing $ \langle objs \rangle$''}. 
Here, \textit{$ \langle \text{scene\_class} \rangle$} corresponds to the scene label of the real image, and \textit{$ \langle objs \rangle$} represents the list of object names, extracted from the object annotations from dataset. The model is configured with guidance scale = 6.5, num\_inference\_steps = 30, and an output resolution of $768 \times 768$ to ensure high-quality synthesis. This approach generates fake images conditioned solely on semantic descriptions. Example outputs are shown in Figure~\ref{fig:output_fakes}(a).

\vspace{-.8em}
%\noindent \textbf{Structure-Guided Fake Image Generation}
% \vspace{-1.8em}
\paragraph{Structure-Guided Fake Image Generation}
In this approach, Stable Diffusion v1.5 is used with the real image as conditioning input along with a text prompt containing the annotated object labels. This version is lightweight, stable, and well-suited
for structure-preserving transformations. This configuration preserves the spatial structure of the original image while introducing controlled generative variation. The model generates a fake image at the same resolution as the input.
The prompt used is:
\textit{``Generate an image in the style of the given image. It should depict a $ \langle \text{scene\_class} \rangle$ scene and must contain: $ \langle \text{objs} \rangle$.''}
Here, \textit{$ \langle \text{scene\_class} \rangle$ }denotes the scene label and \textit{$ \langle \text{objs} \rangle$} denotes the list of annotated objects. The generation uses 30 inference steps, guidance scale = 12.0, and strength = 0.35. The relatively low strength preserves structural layout, while the higher guidance scale enforces adherence to the object list in text prompt, ensuring a fair comparison with real images. Example outputs are shown in Figure~\ref{fig:output_fakes}(b).

% \vspace{-.8em}
\subsection{Compare Neuron Activations}
%\vspace{-.5em}
The generated fake images are passed to the same trained CNN model to maintain consistency in representation extraction. For each fake image, activation values from the 64-neuron final dense layer are recorded under identical model parameters and inference conditions as the real images. These activations are then compared pairwise with their corresponding real images to identify potential representation-level differences.

%\vspace{-1em}
\subsection{Statistical Testing of Activations}
%\vspace{-.5em}
To statistically evaluate activation differences between real and fake image pairs, the Wilcoxon signed-rank test~\citep{10.1093/jee/39.2.269} is utilized which is a non-parametric test designed for paired comparisons. 
For each pair, the test first computes the difference between the real-image activation and the corresponding fake-image activation. It then ranks the absolute non-zero differences and evaluates whether the signed ranks are systematically shifted in one direction. The null hypothesis is that the median paired difference is zero, meaning that real and fake images do not show a systematic activation difference.
In addition, the rank-biserial correlation is used as an effect size measure~\citep{doi:10.2466/11.IT.3.1}, quantifying the magnitude and direction of activation differences between paired samples. 
To complement the p-values and effect sizes, 95\% bootstrap confidence intervals are computed for the paired real-fake differences. Bootstrapping is performed by repeatedly resampling the observed paired differences with replacement and estimating the mean difference for each resampled set. The 2.5\textsuperscript{th} and 97.5\textsuperscript{th} percentiles of the resulting bootstrap distribution are used as the lower and upper bounds of the 95\% confidence interval.

Based on this statistical analysis, four hypotheses and the claim are evaluated using activation values extracted from the 64-neuron last dense layer; therefore, the reported comparisons refer specifically to this representation layer. The hypotheses and claim are:

\noindent \textbf{Hypothesis 1:} Real images exhibit a higher number of activated neurons compared to their fake counterparts.

\noindent \textbf{Hypothesis 2: }Fake images produce lower activation values for relevant confirmed neurons than real images.

\noindent \textbf{Hypothesis 3: }For neurons that activate strongly in real images, fake images have fewer activated neurons than real images.

\noindent \textbf{Hypothesis 4: }For each fake image, the number of activated confirmed neurons is less than its real counterpart.

% \noindent Moreover, the following claim is evaluated:

\noindent \textbf{Claim: }A significant proportion of neurons demonstrate activation suppression in fake images relative to real images.

For Hypothesis 1, the 64 dense-layer activations of each real-fake image pair are compared after zero filtering, where only neuron positions with real\_activation $>$ 0 and fake\_activation $>$ 0 are retained. This reduces the effect of ReLU-induced zero activations, which can otherwise create many tied values in the analysis.
For Hypothesis 2, neuron relevancy is defined in two ways: semantic alignment between a confirmed neuron’s induced concept and the object annotations of the real image, and whether the neuron fires for the real image, i.e., its activation is $\geq$ 80\% of its maximum activation across all real images. Zero filtering is then applied to compare only nonzero activation values.
For Hypothesis 3, a neuron is considered activated if its activation is $\geq$ 80\% of its maximum activation across all real images. For each real-fake pair, neurons activated in the real image are retained, and the analysis tests whether the corresponding fake image activates the same neurons above the threshold.
For Hypothesis 4, the analysis is restricted to confirmed neurons only. For each real-fake pair, the number of confirmed neurons activated in the real image is compared with the corresponding count in the fake image to test whether real images activate significantly more confirmed neurons.
Table~\ref{tab:summary_hypotheses} maps each hypothesis to its neuron subset, filtering criteria, metric tested, and corresponding interpretation.

\begin{table}[ht]
\centering

\caption{Summary of the filtering criteria and interpretation for the four hypotheses.}
\label{tab:summary_hypotheses}

\resizebox{\columnwidth}{!}{%
\begin{tabular}{
|>{\centering\arraybackslash}m{0.14\linewidth}
|>{\centering\arraybackslash}m{0.23\linewidth}
|>{\centering\arraybackslash}m{0.4\linewidth}
|>{\centering\arraybackslash}m{0.23\linewidth}
|>{\centering\arraybackslash}m{0.37\linewidth}|
}
\hline
\textbf{Hypothesis} &
\textbf{Neuron Subset} &
\textbf{Selection Criteria} &
\textbf{Metric Compared} &
\textbf{Interpretation} \\
\hline
H1 &
All &
Activated ($\geq 0.8 \times \max$), zero-filtered &
Count &
Real images trigger broader neural activity than fake \\
\hline
H2 &
Relevant confirmed &
i) Semantically aligned or ii) activated &
Value &
Real images activate relevant neurons more strongly \\
\hline
H3 &
Activated for real &
Activation threshold &
Shared Count &
Fake images fail to preserve real activation patterns \\
\hline
H4 &
Activated \& confirmed &
Confirmed with activation threshold &
Count &
Real images better activate semantically meaningful neurons \\
\hline
\end{tabular}%
}

\end{table}

% \vspace{-1em}
For evaluating the Claim, the difference between real and fake activations for each neuron across paired images is calculated, thereby quantifying the extent to which activations are reduced in fake images relative to real ones.

%\vspace{-1em}
\begin{table}[bht]
\centering
\caption{Statistical evaluation results of concepts of confirmed neurons from concept induction using Mann-Whitney U. \textbf{Bold} rows represent neurons with $p$-value $\geq 0.05$, where \textbf{the null hypothesis cannot be rejected}. The complete table is represented in Figure~\ref{full_neuron_stats} in Appendix~\ref{apd:mwu}.}
\label{neuron_stats}
\resizebox{\columnwidth}{!}{%
\begin{tabular}{
|>{\centering\arraybackslash}p{0.09\linewidth}
|>{\centering\arraybackslash}p{0.28\linewidth}
|>{\centering\arraybackslash}p{0.07\linewidth}
|>{\centering\arraybackslash}p{0.12\linewidth}
|>{\centering\arraybackslash}p{0.12\linewidth}
|>{\centering\arraybackslash}p{0.12\linewidth}
|>{\centering\arraybackslash}p{0.10\linewidth}
|>{\centering\arraybackslash}p{0.15\linewidth}
|>{\centering\arraybackslash}p{0.11\linewidth}
|>{\centering\arraybackslash}p{0.14\linewidth}
|
}
\hline
\textbf{Neuron ID} & \textbf{ECII Concepts} & \textbf{TLA \%} & \textbf{Non-TLA \%} & \textbf{Target Median} & \textbf{Non-Target Median} & \textbf{Target Mean} & \textbf{Non-Target Mean} & \textbf{$z$-score} & \textbf{$p$-value} \\
\hline\hline
0  & snowy\_mountain                 & 95  & 54.44 & 7.05 & 0.25 & 6.12 & 1.04 &  -6.57 & $<0.00001$ \\ \hline
7  & sky\_and\_snowy\_mountain       & 95  & 38.81 & 2.81 & 0.00 & 2.72 & 0.56 &  -5.92 & $<0.00001$ \\ \hline
9  & fence\_and\_central             & 100 & 62.30 & 4.03 & 0.76 & 3.97 & 1.46 &  -5.54 & $<0.00001$ \\ \hline
% 11 & bathtub                         & 100 & 51.98 & 4.79 & 0.05 & 4.69 & 0.75 &  -7.23 & $<0.00001$ \\ \hline
% 12 & coffee\_and\_bouquet            & 95  & 55.16 & 1.37 & 0.20 & 1.60 & 0.78 &  -3.82 & $0.00006$ \\ \hline
% 16 & skyscraper\_and\_building       & 95  & 45.95 & 2.68 & 0.00 & 2.45 & 0.59 &  -5.57 & $<0.00001$ \\ \hline

... & ...        & ...  & ... & ... & ... & ... & ... &  ... & $<0.00001$ \\ \hline
\textbf{41} & \textbf{skyscraper}      & \textbf{80} & \textbf{69.92} & \textbf{1.21} & \textbf{1.03} & \textbf{2.05} & \textbf{1.46} & \textbf{-1.40} & $\mathbf{0.15536}$ \\ \hline

... & ...    & ...  & ... & ... & ... & ... & ... &  ... & $<0.00001$ \\ \hline
61 & cars                            & 95  & 44.37 & 1.40 & 0.00 & 1.15 & 0.50 &  -4.25 & $<0.00001$ \\ \hline
62 & wardrobe\_and\_telephone         & 90  & 57.30 & 1.42 & 0.37 & 1.53 & 0.94 &  -2.98 & $0.00193$ \\ \hline
\end{tabular}
}
\end{table}

% \vspace{-.6em}
\section{Results \& Discussion}
\label{rd}
%\vspace{-.5em}
Concept induction on SUN2012 yields interpretable neuron labels aligned with semantic concepts. Out of the 64 dense-layer neurons, 25 achieve TLA $\geq$ 80\%, and 24 of these also show statistically significant separation between target and non-target activations (Mann-Whitney U, p $<$ 0.05). Table~\ref{neuron_stats} summarizes 
% \pascal{If you need to save space, you could give only a few rows here, and point to a full table in the appendix/supplementary material.} \mss{Done} 
these results, including confirmed labels such as desk, skyscraper, cars, bidet, and bridge. Some labels consist of combined concepts (e.g., ``coffee\_and\_bouquet'') because ECII outputs logical conjunctions of class expressions (e.g., coffee $\sqcap$ bouquet).

%\vspace{-1em}
\begin{table}[b]
\small
\centering
\caption{Hypothesis 1 statistical results with SUN2012 dataset, InceptionV3 CNN model,
and Stable Diffusion image generation model.}
\label{hyp_1_stats}

\resizebox{\columnwidth}{!}{%
\begin{tabular}{
|>{\centering\arraybackslash}m{0.24\linewidth}
|>{\centering\arraybackslash}m{0.45\linewidth}
|>{\centering\arraybackslash}m{0.5\linewidth}|
}
\hline

\textbf{Statistic} &
\textbf{Text-Guided Fake Image Generation} &
\textbf{Structure-Guided Fake Image Generation} \\
\hline

Images paired & 793 & 793 \\
\hline

Nonzero diffs used & 402 & 348 \\
\hline

Median(diff) & 1.000 & 1.000 \\
\hline

Mean(diff) & 0.542 & 0.787 \\
\hline

Prop(real $>$ fake) & 0.600 & 0.681 \\
\hline

$p$-value & $2.749528\times10^{-6}$ & $2.939991\times10^{-12}$ \\
\hline

Effect size $r_{rb}$ & 0.199 & 0.362 \\
\hline

95\% bootstrap CI & [0.33, 0.76] & [0.57, 1.01] \\
\hline

Decision & Reject null hypothesis ($p < 0.05$) &
Reject null hypothesis ($p < 0.05$) \\
\hline

\end{tabular}
}
\end{table}

%\vspace{-1em}
%The statistical testing on compare real and fake 
To compare activations for fake images generated from SUN2012 and Stable Diffusion models for Hypothesis 1, a one-sided Wilcoxon signed-rank test is applied with zero filtering to compare real and fake activations. ``Nonzero diffs used'' denotes the number of paired samples with nonzero differences included in the test. ``Median (real-fake)'' indicates the typical difference in firing neurons per pair, while ``Prop (real $>$ fake)'' represents the proportion of pairs where real images show higher activation counts. The effect size ($r_{rb}$) measures the strength and direction of the tendency for real images to produce higher activations than fake images. The results (represented in Table~\ref{hyp_1_stats}) indicate statistically significant differences in both generation settings, with a 95\% bootstrap confidence intervals of [0.33, 0.76] and [0.57, 1.01]. In the text-guided setup, the test shows p-value of $2.749528\times10^{-6}$ with effect size $r_{rb}$ = 0.199 and 60.0\% of pairs having real $>$ fake. The structure-guided setting demonstrates even stronger separation (p-value = $2.94\times10^{-12}$, $r_{rb}$ = 0.362), with 68.1\% of pairs favoring real images, suggesting stronger and more consistent neuron activation for real images compared to their fake counterparts.

The results for Hypothesis 2, presented in Table~\ref{hyp_2_stat} show strong statistical evidence that fake images produce lower activation values in relevant confirmed neurons compared to real images. Across both relevancy definitions and generation approaches, the Wilcoxon test yields extremely small p-values (p $<$ 0.05) with large effect sizes ($r_{rb}$ ranging from 0.423 to 0.987). The high proportion of cases where real activations exceed fake activations (up to 95.9\%) indicates suppression in fake images.
%\vspace{-1em}
\begin{table}[!h]
\small
\centering
\caption{Hypothesis  2 statistical results with SUN2012 dataset, InceptionV3 CNN model,
and Stable Diffusion image generation model.}
\label{hyp_2_stat}

\resizebox{\columnwidth}{!}{%
\begin{tabular}{
|>{\centering\arraybackslash}m{0.20\linewidth}
|>{\centering\arraybackslash}m{0.3\linewidth}
|>{\centering\arraybackslash}m{0.3\linewidth}
|>{\centering\arraybackslash}m{0.3\linewidth}
|>{\centering\arraybackslash}m{0.3\linewidth}|
}
\hline

\textbf{Statistic} &
\textbf{Text-Guided Fake Generation (Concept Alignment)} &
\textbf{Structure-Guided Fake Generation (Concept Alignment)} &
\textbf{Text-Guided Fake Generation (Activation-Based Relevancy)} &
\textbf{Structure-Guided Fake Generation (Activation-Based Relevancy)} \\
\hline

Images paired & 890 & 885 & 244 & 249 \\
\hline

Nonzero diffs used & 890 & 885 & 244 & 249 \\
\hline

Median (real - fake) & 0.7047 & 1.0429 & 2.5363 & 2.2209 \\
\hline

Mean (real - fake) & 0.7421 & 1.2023 & 2.6552 & 2.5786 \\
\hline

Prop(real - fake) & 0.657 & 0.740 & 0.959 & 0.924 \\
\hline

$p$-value & $4.681\times10^{-28}$ & $5.839\times10^{-66}$ & $4.507\times10^{-41}$ & $2.117\times10^{-40}$ \\
\hline

Effect size ($r_{rb}$) & 0.423 & 0.664 & 0.987 & 0.969 \\
\hline

95\% bootstrap CI & [0.62, 0.87] & [1.08, 1.32] & [2.46, 2.85] & [2.33, 2.83] \\
\hline

Decision & Reject null hypothesis ($p < 0.05$) & 
Reject null hypothesis ($p < 0.05$) & 
Reject null hypothesis ($p < 0.05$) & 
Reject null hypothesis ($p < 0.05$) \\
\hline

\end{tabular}
}
\end{table}

%\vspace{-1em}
\begin{table}[!hbt]
\small
\centering
\caption{Hypothesis 3 statistical results with SUN2012 dataset, InceptionV3 CNN model,
and Stable Diffusion image generation model.}
\label{hyp_3_stats}

\resizebox{\columnwidth}{!}{%
\begin{tabular}{
|>{\centering\arraybackslash}m{0.35\linewidth}
|>{\centering\arraybackslash}m{0.45\linewidth}
|>{\centering\arraybackslash}m{0.5\linewidth}|
}
\hline

\textbf{Statistic} &
\textbf{Text-Guided Fake Image Generation} &
\textbf{Structure-Guided Fake Image Generation} \\
\hline

Images paired & 793 & 793 \\
\hline

Nonzero diffs used & 299 & 264 \\
\hline

Mean(real\_count) & 0.803 & 0.803 \\
\hline

Mean(fake\_count) & 0.057 & 0.158 \\
\hline

Median (real\_count - fake\_count) & 1.000 & 1.000 \\
\hline

Prop(real\_count $>$ fake\_count) & 1.000 & 1.000 \\
\hline

$p$-value & $1.832312\times10^{-53}$ & $2.663170\times10^{-47}$ \\
\hline

Effect size ($r_{rb}$) & 1.000 & 1.000 \\
\hline

95\% bootstrap CI & [1.81, 2.16] & [1.77, 2.12]  \\
\hline

Decision & Reject null hypothesis ($p < 0.05$) &
Reject null hypothesis ($p < 0.05$) \\
\hline

\end{tabular}
}
\end{table}

For Hypothesis 3, the results in Table~\ref{hyp_3_stats} show that fake images activate fewer high-threshold neurons among those strongly activated by real images, where activation is defined as $\geq$ 80\% of the neuron’s maximum activation across real images. In both generation approaches, the median difference (real\_count - fake\_count) is 1.000. The Wilcoxon signed-rank test gives highly significant p-values (1.83 $\times$ 10\textsuperscript{-53} and 2.66 $\times$ 10\textsuperscript{-47}), with $r_{rb}$ = 1.000, indicating suppression of high-threshold neuron activations in fake images.

%\vspace{-1em}
\begin{table}[!hbt]
\small
\centering
\caption{Hypothesis 4 statistical results  with SUN2012 dataset, InceptionV3 CNN model,
and Stable Diffusion image generation model.}
\label{hyp_4_stats}

\resizebox{\columnwidth}{!}{%
\begin{tabular}{
|>{\centering\arraybackslash}m{0.26\linewidth}
|>{\centering\arraybackslash}m{0.49\linewidth}
|>{\centering\arraybackslash}m{0.49\linewidth}|
}
\hline

\textbf{Statistic} &
\textbf{Text-Guided Fake Image Generation} &
\textbf{Structure-Guided Fake Image Generation} \\
\hline

Images paired & 793 & 793 \\
\hline

Nonzero diffs used & 212 & 189 \\
\hline

Median (real-fake) counts & 1.000 & 1.000 \\
\hline

Prop(real $>$ fake) & 0.646 & 0.709 \\
\hline

$p$-value & $6.272851\times10^{-6}$ & $1.724670\times10^{-8}$ \\
\hline

Effect size ($r_{rb})$ & 0.292 & 0.418 \\
\hline

95\% bootstrap CI & [0.29, 0.7] & [0.46, 0.88] \\
\hline

Decision & Reject null hypothesis ($p < 0.05$) &
Reject null hypothesis ($p < 0.05$) \\
\hline

\end{tabular}
}
\end{table}

\begin{figure}[!htb]
    \centering
    \includegraphics[width=0.8\textwidth]{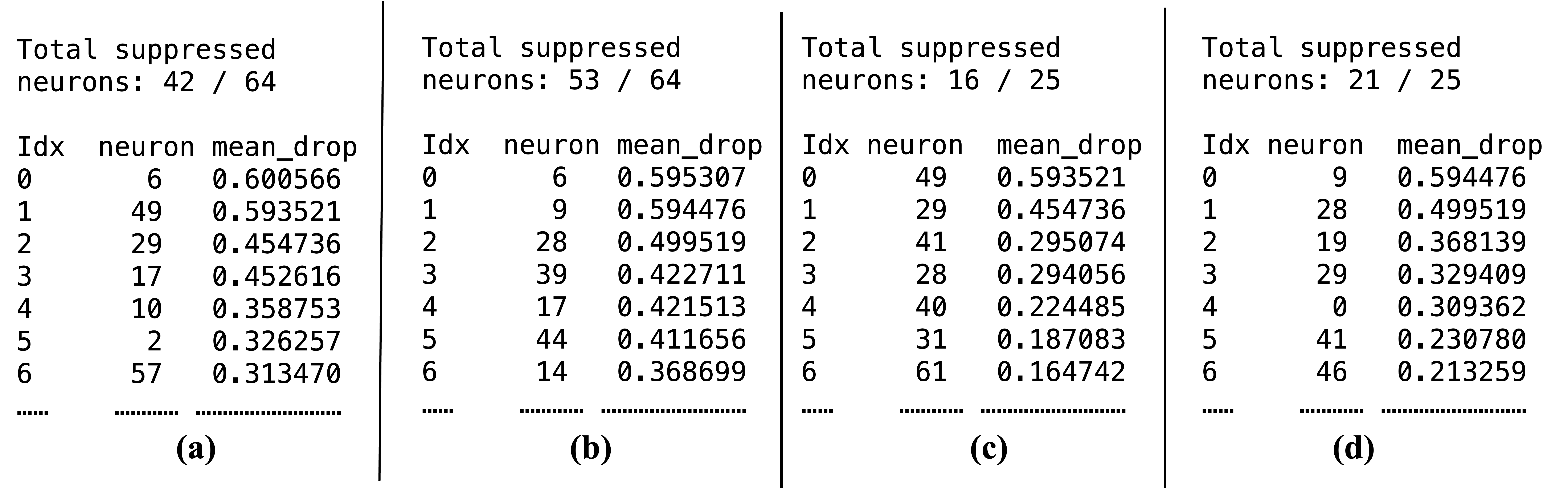}
    \caption{
   %  \eyv{Could you provide the text instead of the image? I can make it into a table.} 
   % \mss{I tried adding this in a table, but it covers a lot space than the image, eventually the paper exceeds pg limit (10).} \eyv{I meant, could you give \emph{me} the text and I would try to make a table. But it doesn't look like I'll have time now so no worries.} 
   Suppression in neurons and their mean activation drop for SUN2012 dataset, InceptionV3 CNN model,
and Stable Diffusion image generation model. (a) Text-guided fake image generation (all neurons), (b) Structure-guided fake image generation (all neurons), (c) Text-guided generation for confirmed neurons, and (d) Structure-guided generation for confirmed neurons. Complete table is represented in Appendix \ref{claim_sun_stab}.}
    \label{fig:claim_sun_stab}
\end{figure}

For Hypothesis 4, the results in Table~\ref{hyp_4_stats} provide statistically significant evidence that fake images activate fewer confirmed neurons than their real counterparts. In the text-guided setting, 212 nonzero pairs show a median difference of 1 neuron, with 64.6\% of cases favoring real images (p $\approx$ 6.27 $\times$ 10\textsuperscript{-6}, $r_{rb}$ = 0.292), and a confidence interval of [0.29, 0.7]. The structure-guided setting shows a stronger effect, with 70.9\% of pairs favoring real images (p $\approx$ 1.72 $\times$ 10\textsuperscript{-8}, $r_{rb}$ = 0.418). These findings support that confirmed semantic neurons fire more frequently for real images.

Evaluation of the Claim shows significant activation suppression in fake images: 42 of 64 neurons are suppressed under text-guided generation and 53 of 64 under structure-guided generation. Among the 25 confirmed neurons, suppression is observed in 16 (text-guided) and 21 (structure-guided). The results are shown in Figure~\ref{fig:claim_sun_stab}.

%\vspace{-1em}
% \begin{figure}[!ht]
%     \centering
%     \includegraphics[width=7cm, height=11cm]{Figures/suppress_claim.png}
%     \caption{Suppressed neurons and their mean activation drop under - (a) text-guided and (b) structure-guided fake image generation.}
%     \label{fig:Claim1}
% \end{figure}

% %\vspace{-.5em}
% \begin{figure}[!htb]
%     \centering
%     \includegraphics[width=9cm, height=7cm]{Figures/suppress_claim_confirmed.png}
%     \caption{Suppression in confirmed neurons and their mean activation drop under - (a) text-guided and (b) structure-guided fake image generation.}
%     \label{fig:Claim2}
% \end{figure}

% Overall, the results from all four hypotheses and the suppression claim support the main research hypothesis that 
% % \mss{changed CNNs from DNNs (it was overstated)}
% CNNs internally represent real and fake images differently. Even with matched semantic content, fake images produce weaker and altered hidden-layer activation patterns, indicating measurable differences in internal representations. 
% The nonzero-difference counts further show that these effects vary across image pairs and are stronger in some comparisons than others.
Overall, for the SUN2012-InceptionV3 setting with Stable Diffusion-generated images, the results support the main research hypothesis that real and fake images can produce different CNN hidden-layer activation patterns. The nonzero-difference counts further indicate that these effects vary across image pairs and are stronger in some comparisons than others.

\begin{figure}[tb]
    \centering
    \includegraphics[height=7cm,width=7cm]{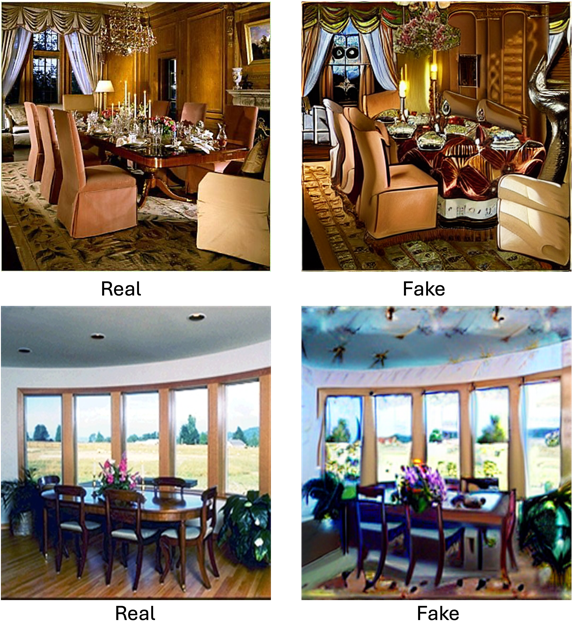}
    \caption{Structural Distortion Effects in Structure-Guided
Fake Image Generation}
    \label{fig:cons}
\end{figure}

However, in structure-guided generation, structural degradation introduced by the image generation model, as illustrated in Figure~\ref{fig:cons}, may reduce neuron activations in fake images. This effect is more noticeable for low-resolution inputs and less pronounced for high-resolution images. In contrast, text-guided generation does not show the same input-dependent distortion. Further hyperparameter refinement or alternative generative models may help generate higher-quality fake images with better structural consistency.

%\vspace{-.5em}

% \vspace{-1em}
To examine whether activation differences are caused by image quality degradation, JPEG compression and Gaussian blur are applied to real images, and the same hypothesis-testing procedure is repeated by comparing each real image with its transformed counterpart (results in Appendix~\ref{apd:degradation}). The results show that degradation can contribute to activation changes, especially under stronger blur, but JPEG compression at q = 70 (medium) and mild blur (k = 3, $\sigma$ = 1) do not fully reproduce the patterns observed in the real-fake comparison. Thus, image quality degradation remains a potential confounding factor, particularly when degradation is strong. 
% The data underlying this image degradation analysis are provided in Appendix~\ref{apd:degradation}.

To examine whether the observed real-fake activation differences are specific to the initial SUN2012, InceptionV3, and Stable Diffusion configuration, additional experiments are conducted using ResNet50V2, ADE20K~\citep{10.1007/s11263-018-1140-0} with the 10 largest scene categories, and FLUX.1 [dev] as an additional image generation model. 
% FLUX.1 [dev] is a recent open-weight image generation model from Black Forest Labs. It is a 12B-parameter rectified flow transformer and is trained using guidance distillation, which is intended to improve inference efficiency while maintaining strong image quality and prompt-following ability.
The choice of ResNet50V2 and ADE20K is motivated by prior work~\citep{10.1007/978-3-031-71170-1_12} on the Concept Induction-Based Neuron Interpretability Framework. Therefore, this setting allows to evaluate the proposed real-fake activation analysis in a configuration that is both different from the initial experiment and grounded in an existing neuron-interpretability setup. The findings depicted in Appendix~\ref{apd:second} suggest that generated images show distinguishable hidden-neuron activation behavior compared with real images in this setup. However, the strength and form of these differences may vary across datasets, CNN architectures, and image generation models. A broader investigation of such variations is left for future work.

%\vspace{-1em}
\section{Conclusion}
%\vspace{-.5em}
% In this work, consistent evidence demonstrates clear differences in hidden-layer activation patterns between real and fake images, even when semantic content is matched. Across all four hypotheses and the suppression analysis (Claim), fake images exhibit weaker and less stable neuron activations, revealing a significant internal signal that distinguishes real from generated images. This signal suggests that hidden-neuron behavior may provide a principled basis for realistic deepfake detection. Future work should focus on systematically leveraging this activation-based signal for practical fake-image detection, improving generative quality through hyperparameter refinement or alternative models to better separate structural degradation from deeper representational differences, and evaluating adversarial robustness to assess whether activation-based detection remains effective when generative models attempt to minimize internal activation disparities.

% \mss{modified conclusion  reflecting the additional experiments and comment of Reviewer 2}

In this work, real and fake images are observed to evoke different hidden-layer activation behavior, even when their semantic content is matched. Across the proposed hypotheses, fake images often produce weaker or less stable neuron activations than their real counterparts, suggesting that hidden-neuron activations may provide useful signals for studying real–fake image differences. Degradation experiments show that JPEG compression and Gaussian blur can affect neuron activations, especially under stronger degradation. However, these transformations did not fully reproduce the real-fake patterns, suggesting that degradation may contribute to the observed differences but is unlikely to be the only factor. Experiments with different CNN architectures, datasets, Stable Diffusion-based settings, and FLUX.1 further support the claim that fake images can evoke different hidden-layer activation patterns across multiple settings.
Overall, hidden-layer activation analysis appears to be a useful direction for studying real–fake image differences. However, the findings remain an initial step rather than a complete generalization across all image-generation approaches or full invariance to all kinds of image-quality degradation. Future work should examine broader datasets, architectures, image domains, and generative models, while also studying why activation differences occur, such as by grouping induced neuron concepts into higher-level semantic categories. Another important direction is to investigate potential distribution shifts introduced by prompt generation or generative-model biases, such as systematic differences in style, lighting, composition, or scene complexity between real and fake images. Future work should also evaluate adversarial robustness when generative models attempt to minimize internal activation disparities.
In addition, the current findings further indicate that it is reasonable to investigate whether hidden-neuron activation analysis can be used to improve existing fake image detection approaches, which will be explored in future work.

%\section{Citations and Bibliography}
%\label{sec:cite}
%
%The \textsf{jmlr} class automatically loads \textsf{natbib}.
%This sample file has the citations defined in the accompanying
%BibTeX file \texttt{pmlr-sample.bib}. For a parenthetical
%citation use \verb|\citep|. For example
%\citep{guyon-elisseeff-03}. For a textual citation use
%\verb|\citet|. For example~\citet{guyon2007causalreport}.
%Both commands may take a comma-separated list, for example
%\citet{guyon-elisseeff-03,guyon2007causalreport}.
%
%These commands have optional arguments and have a starred
%version. See the \textsf{natbib} documentation for further
%details.\footnote{Either \texttt{texdoc natbib} or
%\url{http://www.ctan.org/pkg/natbib}}
%
%The bibliography is displayed using \verb|\bibliography|.
\section*{Acknowledgments}
The authors acknowledge partial funding under Kansas State University’s Game Changing Research Initiative (GRIP) program.
% \vspace{-.5em}
\section*{Declaration on Generative AI}
%For this work,
The authors used ChatGPT-5 for grammar and spelling checks. Subsequently, the authors reviewed and edited the text as needed, and take full responsibility for the content. 

\bibliography{nesy2026-ref}

@misc{flux2024,
    author={{Black Forest Labs}},
    title={Official inference repo for {FLUX.1} models},
    year={2025},
    howpublished={\url{https://github.com/black-forest-labs/flux}},
}

@article{10.1007/s11263-018-1140-0,
author = {Zhou, Bolei and Zhao, Hang and Puig, Xavier and Xiao, Tete and Fidler, Sanja and Barriuso, Adela and Torralba, Antonio},
title = {Semantic Understanding of Scenes Through the {ADE20K} Dataset},
year = {2019},
issue_date = {March 2019},
publisher = {Kluwer Academic Publishers},
address = {USA},
volume = {127},
number = {3},
issn = {0920-5691},
url = {https://doi.org/10.1007/s11263-018-1140-0},
doi = {10.1007/s11263-018-1140-0},
journal = {Int. J. Comput. Vision},
month = mar,
pages = {302–321},
numpages = {20}
}

@article{G2024DetectingAI,
  title={Detecting {AI}-generated images with {CNN} and Interpretation using Explainable {AI}},
  author={Bharathi Mohan G. and Prasanna Kumar Rangarajan and Akilesh rao S and Mandava Sukesh and Abinandhini D M and Jaikanth Y},
  journal={2024 IEEE International Conference on Contemporary Computing and Communications (InC4)},
  year={2024},
  volume={1},
  pages={1-6},
  url={https://api.semanticscholar.org/CorpusID:272373462}
}

@article{HasanAbir2023DetectingDI,
  title={Detecting Deepfake Images Using Deep Learning Techniques and Explainable {AI} Methods},
  author={Wahidul Hasan Abir and Faria Rahman Khanam and Kazi Nabiul Alam and Myriam Hadjouni and Hela Elmannai and Sami Bourouis and Rajesh Dey and Mohammad Monirujjaman Khan},
  journal={Intell. Autom. Soft Comput.},
  year={2023},
  volume={35},
  pages={2151-2169},
  url={https://api.semanticscholar.org/CorpusID:250720271}
}

@article{Bird2023CIFAKEIC,
  title={{CIFAKE}: Image Classification and Explainable Identification of {AI}-Generated Synthetic Images},
  author={Jordan J. Bird and Ahmad Lotfi},
  journal={IEEE Access},
  year={2023},
  volume={12},
  pages={15642-15650},
  url={https://api.semanticscholar.org/CorpusID:257757303}
}

@article{Malolan2020ExplainableDD,
  title={Explainable Deep-Fake Detection Using Visual Interpretability Methods},
  author={Badhrinarayan Malolan and Ankit Parekh and Faruk Kazi},
  journal={2020 3rd International Conference on Information and Computer Technologies (ICICT)},
  year={2020},
  pages={289-293},
  url={https://api.semanticscholar.org/CorpusID:218650902}
}

@ARTICLE{11316647,
  author={Aleem, Muhammad and Umair, Muhammad and Zubair, Muhammad and Ibrahim, Rozeena and Naseem, Muhammad Tahir and Raza, Muhammad Mohsin and Ali, Muhammad Nadeem and Kim, Byung-Seo},
  journal={IEEE Access}, 
  title={Seeing Through the Fake: Explainable {AI} With Multiple {CNNs} for Deepfake Detection}, 
  year={2026},
  volume={14},
  number={},
  pages={131-162},
  doi={10.1109/ACCESS.2025.3649128},
}

@inproceedings{10.1145/3643491.3660292,
author = {Tsigos, Konstantinos and Apostolidis, Evlampios and Baxevanakis, Spyridon and Papadopoulos, Symeon and Mezaris, Vasileios},
title = {Towards Quantitative Evaluation of Explainable {AI} Methods for Deepfake Detection},
year = {2024},
isbn = {9798400705526},
publisher = {Association for Computing Machinery},
address = {New York, NY, USA},
xurl = {https://doi.org/10.1145/3643491.3660292},
doi = {10.1145/3643491.3660292},
booktitle = {Proceedings of the 3rd ACM International Workshop on Multimedia AI against Disinformation},
pages = {37–45},
numpages = {9},
location = {Phuket, Thailand},
series = {MAD '24}
}

@article{Li2024FakeBenchPE,
  title={{FakeBench}: Probing Explainable Fake Image Detection via Large Multimodal Models},
  author={Yixuan Li and Xuelin Liu and Xiaoyang Wang and Bu Sung Lee and Shiqi Wang and Anderson Rocha and Weisi Lin},
  journal={IEEE Transactions on Information Forensics and Security},
  year={2024},
  volume={20},
  pages={8730-8745},
  url={https://api.semanticscholar.org/CorpusID:272525421}
}

@inproceedings{10.1007/978-3-031-71170-1_12,
author = {Dalal, Abhilekha and Rayan, Rushrukh and Barua, Adrita and Vasserman, Eugene Y. and Sarker, Md Kamruzzaman and Hitzler, Pascal},
title = {On the Value of Labeled Data and Symbolic Methods for Hidden Neuron Activation Analysis},
year = {2024},
isbn = {978-3-031-71169-5},
publisher = {Springer-Verlag},
address = {Berlin, Heidelberg},
xurl = {https://doi.org/10.1007/978-3-031-71170-1_12},
doi = {10.1007/978-3-031-71170-1_12},
booktitle = {Neural-Symbolic Learning and Reasoning: 18th International Conference, NeSy 2024, Barcelona, Spain, September 9–12, 2024, Proceedings, Part II},
pages = {109–131},
numpages = {23},
location = {Barcelona, Spain}
}

@misc{sarma2026casestudyconceptinduction,
      title={A Case Study on Concept Induction for Neuron-Level Interpretability in CNN}, 
      author={Moumita Sen Sarma and Samatha Ereshi Akkamahadevi and Pascal Hitzler},
      year={2026},
      eprint={2603.00197},
      archivePrefix={arXiv},
      primaryClass={cs.CV},
      url={https://arxiv.org/abs/2603.00197}, 
}

@InProceedings{10.1007/978-3-030-65384-2_6,
author="Sarker, Md Kamruzzaman
and Schwartz, Joshua
and Hitzler, Pascal
and Zhou, Lu
and Nadella, Srikanth
and Minnery, Brandon
and Juvina, Ion
and Raymer, Michael L.
and Aue, William R.",
editor="Villaz{\'o}n-Terrazas, Boris
and Ortiz-Rodr{\'i}guez, Fernando
and Tiwari, Sanju M.
and Shandilya, Shishir K.",
title="Wikipedia Knowledge Graph for Explainable AI",
booktitle="Knowledge Graphs and Semantic Web",
year="2020",
publisher="Springer International Publishing",
address="Cham",
pages="72--87",
isbn="978-3-030-65384-2"
}

@article{Sarker_Hitzler_2019,
title={Efficient Concept Induction for Description Logics},
volume={33},
xurl={https://ojs.aaai.org/index.php/AAAI/article/view/4161},
DOI={10.1609/aaai.v33i01.33013036},
number={01},
journal={Proceedings of the AAAI Conference on Artificial Intelligence},
author={Sarker, Md Kamruzzaman and Hitzler, Pascal},
year={2019},
month={Jul.},
pages={3036-3043}
}

@article{ritika_diffusion,
    title = {Mastering Diffusion Models: A Guide to Image Generation with Stable Diffusion},
    author = {Ritika},
    month = {September},
    day = {29},
    year = {2023},
    journal = {Analytics Vidhya},
    url = {https://www.analyticsvidhya.com/blog/2023/09/image-generation-using-stable-diffusion/},
    note = {Accessed 1-1-2026}
}

@article{10.1093/jee/39.2.269,
    author = {Wilcoxon, Frank},
    title = {Individual Comparisons of Grouped Data by Ranking Methods},
    journal = {Journal of Economic Entomology},
    volume = {39},
    number = {2},
    pages = {269-270},
    year = {1946},
    month = {04},
    issn = {0022-0493},
    doi = {10.1093/jee/39.2.269},
    url = {https://doi.org/10.1093/jee/39.2.269},
    eprint = {https://academic.oup.com/jee/article-pdf/39/2/269/19198583/jee39-0269.pdf},
}

@article{doi:10.1142/S0129065725500522,
author = {Rojas-Albarrac\'{\i}n, Gabriel and Pereira, Ant\'{o}nio and Fern\'{a}ndez-Caballero, Antonio and L\'{o}pez, Mar\'{\i}a T.},
title = {Expanding Domain-Specific Datasets with Stable Diffusion Generative Models for Simulating Myocardial Infarction},
journal = {International Journal of Neural Systems},
volume = {35},
number = {10},
pages = {2550052},
year = {2025},
doi = {10.1142/S0129065725500522},
    note ={PMID: 40760711},

URL = { https://doi.org/10.1142/S0129065725500522},
eprint = {https://doi.org/10.1142/S0129065725500522}

}

@ARTICLE{11079597,
  author={Raghavendra, S. and Vijayalakshmi and Vainidhi and Abhilash, S. K. and Madhav Nookala, Venu and Arun Kumar, P. V. and Ramyashree},
  journal={IEEE Access}, 
  title={SVPDSA: Selective View Perception Data Synthesis With Annotations Using Lightweight Diffusion Network}, 
  year={2025},
  volume={13},
  number={},
  pages={124051-124067},
  doi={10.1109/ACCESS.2025.3588542}}

@inproceedings{NEURIPS2023_505df5ea,
 author = {Lu, Zeyu and Huang, Di and BAI, LEI and Qu, Jingjing and Wu, Chengyue and Liu, Xihui and Ouyang, Wanli},
 booktitle = {Advances in Neural Information Processing Systems},
 editor = {A. Oh and T. Naumann and A. Globerson and K. Saenko and M. Hardt and S. Levine},
 pages = {25435--25447},
 publisher = {Curran Associates, Inc.},
 title = {Seeing is not always believing: Benchmarking Human and Model Perception of AI-Generated Images},
 url = {https://proceedings.neurips.cc/paper_files/paper/2023/file/505df5ea30f630661074145149274af0-Paper-Datasets_and_Benchmarks.pdf},
 volume = {36},
 year = {2023}
}

@ARTICLE{9115874,
  author={Verdoliva, Luisa},
  journal={IEEE Journal of Selected Topics in Signal Processing}, 
  title={Media Forensics and DeepFakes: An Overview}, 
  year={2020},
  volume={14},
  number={5},
  pages={910-932},
  doi={10.1109/JSTSP.2020.3002101}}

@Article{app131910980,
AUTHOR = {Sharma, Dilip Kumar and Singh, Bhuvanesh and Agarwal, Saurabh and Garg, Lalit and Kim, Cheonshik and Jung, Ki-Hyun},
TITLE = {A Survey of Detection and Mitigation for Fake Images on Social Media Platforms},
JOURNAL = {Applied Sciences},
VOLUME = {13},
YEAR = {2023},
NUMBER = {19},
ARTICLE-NUMBER = {10980},
URL = {https://www.mdpi.com/2076-3417/13/19/10980},
ISSN = {2076-3417},
DOI = {10.3390/app131910980}
}

@article{article,
author = {Gupta, Divya},
year = {2024},
month = {06},
pages = {45-56},
title = {Generative AI and Deep fake s: Ethical Implications and Detection Techniques},
volume = {1},
journal = {Journal of Science, Technology and Engineering Research},
doi = {10.64206/21rgkc40}
}

@article{
doi:10.1073/pnas.2120481119,
author = {Sophie J. Nightingale  and Hany Farid },
title = {AI-synthesized faces are indistinguishable from real faces and more trustworthy},
journal = {Proceedings of the National Academy of Sciences},
volume = {119},
number = {8},
pages = {e2120481119},
year = {2022},
doi = {10.1073/pnas.2120481119},
URL = {https://www.pnas.org/doi/abs/10.1073/pnas.2120481119},
eprint = {https://www.pnas.org/doi/pdf/10.1073/pnas.2120481119},
}

@Article{jimaging8110310,
AUTHOR = {Man, Keith and Chahl, Javaan},
TITLE = {A Review of Synthetic Image Data and Its Use in Computer Vision},
JOURNAL = {Journal of Imaging},
VOLUME = {8},
YEAR = {2022},
NUMBER = {11},
ARTICLE-NUMBER = {310},
URL = {https://www.mdpi.com/2313-433X/8/11/310},
PubMedID = {36422059},
ISSN = {2313-433X},
}

@article{Zhao2024ARO,
  title={A review of convolutional neural networks in computer vision},
  author={Xia Zhao and Limin Wang and Yufei Zhang and Xuming Han and Muhammet Deveci and Milan Deepak Parmar},
  journal={Artificial Intelligence Review},
  year={2024},
  volume={57},
  url={https://api.semanticscholar.org/CorpusID:268673911}
}

@article{HAAR2023105606,
title = {An analysis of explainability methods for convolutional neural networks},
journal = {Engineering Applications of Artificial Intelligence},
volume = {117},
pages = {105606},
year = {2023},
issn = {0952-1976},
doi = {https://doi.org/10.1016/j.engappai.2022.105606},
url = {https://www.sciencedirect.com/science/article/pii/S0952197622005966},
author = {Lynn Vonder Haar and Timothy Elvira and Omar Ochoa},
}

@INPROCEEDINGS{8237336,
  author={Selvaraju, Ramprasaath R. and Cogswell, Michael and Das, Abhishek and Vedantam, Ramakrishna and Parikh, Devi and Batra, Dhruv},
  booktitle={2017 IEEE International Conference on Computer Vision (ICCV)}, 
  title={Grad-CAM: Visual Explanations from Deep Networks via Gradient-Based Localization}, 
  year={2017},
  volume={},
  number={},
  pages={618-626},
  doi={10.1109/ICCV.2017.74}}

@misc{
zheng2022shapcam,
title={Shap-{CAM}: Visual Explanations for Convolutional Neural Networks based on Shapley Value},
author={Quan Zheng and Ziwei Wang and Jiwen Lu and Jie Zhou},
year={2022},
url={https://openreview.net/forum?id=C1lXY_T1LTs}
}

@INPROCEEDINGS{9878449,
  author={Rombach, Robin and Blattmann, Andreas and Lorenz, Dominik and Esser, Patrick and Ommer, Björn},
  booktitle={2022 IEEE/CVF Conference on Computer Vision and Pattern Recognition (CVPR)}, 
  title={High-Resolution Image Synthesis with Latent Diffusion Models}, 
  year={2022},
  volume={},
  number={},
  pages={10674-10685},
  doi={10.1109/CVPR52688.2022.01042}}

@misc{podell2023sdxlimprovinglatentdiffusion,
      title={SDXL: Improving Latent Diffusion Models for High-Resolution Image Synthesis}, 
      author={Dustin Podell and Zion English and Kyle Lacey and Andreas Blattmann and Tim Dockhorn and Jonas Müller and Joe Penna and Robin Rombach},
      year={2023},
      eprint={2307.01952},
      archivePrefix={arXiv},
      primaryClass={cs.CV},
      url={https://arxiv.org/abs/2307.01952}, 
}

@INPROCEEDINGS{10625113,
  author={Jadhav, Balasaheb and Jain, Manas and Jajoo, Adhip and Kadam, Devika and Kadam, Harshvardhan and Kakkad, Toshish},
  booktitle={2024 2nd International Conference on Sustainable Computing and Smart Systems (ICSCSS)}, 
  title={Imagination Made Real: Stable Diffusion for High-Fidelity Text-to-Image Tasks}, 
  year={2024},
  volume={},
  number={},
  pages={773-779},
  doi={10.1109/ICSCSS60660.2024.10625113}}

@article{doi:10.2466/11.IT.3.1,
author = {Dave S. Kerby},
title ={The Simple Difference Formula: An Approach to Teaching Nonparametric Correlation1},
journal = {Comprehensive Psychology},
volume = {3},
number = {},
pages = {11.IT.3.1},
year = {2014},
doi = {10.2466/11.IT.3.1},

URL = { https://journals.sagepub.com/doi/abs/10.2466/11.IT.3.1},
eprint = {https://journals.sagepub.com/doi/pdf/10.2466/11.IT.3.1}

}

@inproceedings{ribeiro2016should,
  title={{``Why Should I Trust You?'' Explaining} the Predictions of Any Classifier},
  author={Ribeiro, Marco Tulio and Singh, Sameer and Guestrin, Carlos},
  booktitle={Proceedings of the 22nd {ACM SIGKDD} International Conference on Knowledge Discovery and Data Mining},
  pages={1135--1144},
  year={2016},
  publisher={ACM},
  doi={10.1145/2939672.2939778}
}

@inproceedings{NIPS2017_7062,
  title={A unified approach to interpreting model predictions},
  author={Lundberg, Scott M and Lee, Su-In},
  booktitle={Advances in {Neural Information Processing Systems} (NeurIPS)},
  volume={30},
  pages={4765--4774},
  year={2017},
  publisher={Curran Associates, Inc.},
 
}

@article{Bach2015OnPE,
  title={On Pixel-Wise Explanations for Non-Linear Classifier Decisions by Layer-Wise Relevance Propagation},
  author={Sebastian Bach and Alexander Binder and Gr{\'e}goire Montavon and Frederick Klauschen and Klaus-Robert M{\"u}ller and Wojciech Samek},
  journal={PLoS ONE},
  year={2015},
  volume={10},
  url={https://api.semanticscholar.org/CorpusID:9327892}
}
\clearpage
\appendix
%\vspace{-.5em}
%\section{First Appendix}\label{apd:first}
\section{Statistical Evaluation Results of Concept Induction using Mann-Whitney U}\label{apd:mwu}

\begin{table}[!ht]

\centering
\caption{Statistical evaluation results of concepts of confirmed neurons from concept induction. \textbf{Bold} rows represent neurons with $p$-value $\geq 0.05$, where \textbf{the null hypothesis cannot be rejected}.}
\label{full_neuron_stats}
\resizebox{\columnwidth}{!}{%
\begin{tabular}{
|>{\centering\arraybackslash}p{0.09\linewidth}
|>{\centering\arraybackslash}p{0.28\linewidth}
|>{\centering\arraybackslash}p{0.07\linewidth}
|>{\centering\arraybackslash}p{0.12\linewidth}
|>{\centering\arraybackslash}p{0.12\linewidth}
|>{\centering\arraybackslash}p{0.12\linewidth}
|>{\centering\arraybackslash}p{0.10\linewidth}
|>{\centering\arraybackslash}p{0.15\linewidth}
|>{\centering\arraybackslash}p{0.11\linewidth}
|>{\centering\arraybackslash}p{0.14\linewidth}
|
}
\hline
\textbf{Neuron ID} & \textbf{ECII Concepts} & \textbf{TLA \%} & \textbf{Non-TLA \%} & \textbf{Target Median} & \textbf{Non-Target Median} & \textbf{Target Mean} & \textbf{Non-Target Mean} & \textbf{$z$-score} & \textbf{$p$-value} \\
\hline\hline
0  & snowy\_mountain                 & 95  & 54.44 & 7.05 & 0.25 & 6.12 & 1.04 &  -6.57 & $<0.00001$ \\ \hline
7  & sky\_and\_snowy\_mountain       & 95  & 38.81 & 2.81 & 0.00 & 2.72 & 0.56 &  -5.92 & $<0.00001$ \\ \hline
9  & fence\_and\_central             & 100 & 62.30 & 4.03 & 0.76 & 3.97 & 1.46 &  -5.54 & $<0.00001$ \\ \hline
11 & bathtub                         & 100 & 51.98 & 4.79 & 0.05 & 4.69 & 0.75 &  -7.23 & $<0.00001$ \\ \hline
12 & coffee\_and\_bouquet            & 95  & 55.16 & 1.37 & 0.20 & 1.60 & 0.78 &  -3.82 & $0.00006$ \\ \hline
16 & skyscraper\_and\_building       & 95  & 45.95 & 2.68 & 0.00 & 2.45 & 0.59 &  -5.57 & $<0.00001$ \\ \hline
19 & snowy\_mountain                 & 100 & 41.03 & 6.17 & 0.00 & 5.49 & 0.66 &  -7.16 & $<0.00001$ \\ \hline
20 & cars                            & 85  & 44.05 & 1.12 & 0.00 & 1.03 & 0.45 &  -3.78 & $0.00003$ \\ \hline
27 & toilet                          & 100 & 60.71 & 4.44 & 0.53 & 4.04 & 1.13 &  -6.28 & $<0.00001$ \\ \hline
28 & car                             & 100 & 66.43 & 1.84 & 0.81 & 1.86 & 1.37 &  -2.68 & $0.00627$ \\ \hline
29 & bidet                           & 95  & 39.44 & 2.87 & 0.00 & 3.10 & 0.66 &  -6.05 & $<0.00001$ \\ \hline
31 & bidet                           & 90  & 62.94 & 3.37 & 0.64 & 3.05 & 1.22 &  -4.22 & $0.00002$ \\ \hline
35 & bridge                          & 90  & 38.57 & 1.82 & 0.00 & 1.64 & 0.51 &  -4.68 & $<0.00001$ \\ \hline
36 & snowy\_mountain                 & 100 & 66.59 & 5.35 & 0.83 & 5.34 & 1.37 &  -7.17 & $<0.00001$ \\ \hline
40 & sideboard\_and\_soupdish        & 95  & 46.35 & 2.16 & 0.00 & 2.16 & 0.66 &  -5.51 & $<0.00001$ \\ \hline
\textbf{41} & \textbf{skyscraper}      & \textbf{80} & \textbf{69.92} & \textbf{1.21} & \textbf{1.03} & \textbf{2.05} & \textbf{1.46} & \textbf{-1.40} & $\mathbf{0.15536}$ \\ \hline
42 & skyscraper\_and\_building       & 70  & 59.92 & 2.70 & 0.64 & 2.54 & 1.30 &  -2.20 & $0.02322$ \\ \hline
43 & skyscraper                      & 100 & 65.08 & 3.87 & 0.68 & 3.87 & 1.09 &  -6.51 & $<0.00001$ \\ \hline
46 & skyscraper                      & 90  & 61.51 & 2.46 & 0.49 & 2.13 & 1.12 &  -3.08 & $0.00153$ \\ \hline
48 & toilet                          & 100 & 65.56 & 3.53 & 0.66 & 3.70 & 1.30 &  -5.35 & $<0.00001$ \\ \hline
49 & shower\_curtain & 100 & 62.14 & 4.19 & 0.57 & 4.08 & 1.06 &  -6.78 & $<0.00001$ \\ \hline
58 & desk                            & 100 & 52.46 & 1.82 & 0.12 & 1.92 & 1.05 &  -3.92 & $0.00004$ \\ \hline
60 & fruit\_bowl                      & 85  & 37.94 & 1.41 & 0.00 & 1.40 & 0.52 &  -4.20 & $<0.00001$ \\ \hline
61 & cars                            & 95  & 44.37 & 1.40 & 0.00 & 1.15 & 0.50 &  -4.25 & $<0.00001$ \\ \hline
62 & wardrobe\_and\_telephone         & 90  & 57.30 & 1.42 & 0.37 & 1.53 & 0.94 &  -2.98 & $0.00193$ \\ \hline
\end{tabular}
}
\end{table}

\clearpage
\section{Evaluation result of the Claim on SUN2012 dataset, InceptionV3 CNN model, and Stable Diffusion image generation model}
\label{claim_sun_stab}

\begin{figure}[!htb]
    \centering
    \includegraphics[width=15cm, height=16cm]{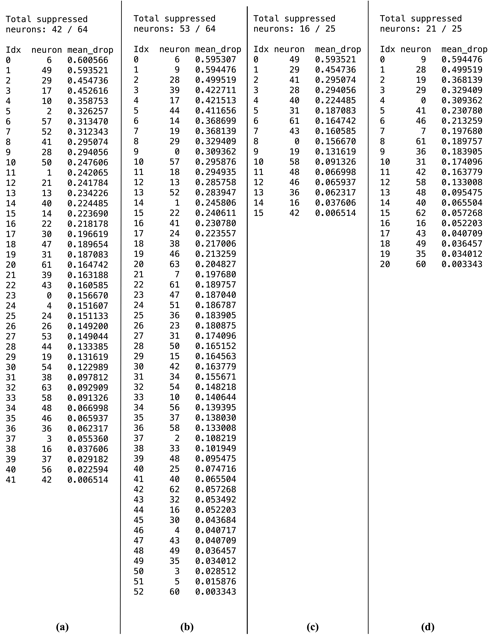}
    \caption{
   %  \eyv{Could you provide the text instead of the image? I can make it into a table.} 
   % \mss{I tried adding this in a table, but it covers a lot space than the image, eventually the paper exceeds pg limit (10).} \eyv{I meant, could you give \emph{me} the text and I would try to make a table. But it doesn't look like I'll have time now so no worries.} 
   Suppression in neurons and their mean activation drop for SUN2012 dataset, InceptionV3 CNN model,
and Stable Diffusion image generation model. (a) Text-guided fake image generation (all neurons), (b) Structure-guided fake image generation (all neurons), (c) Text-guided generation for confirmed neurons, and (d) Structure-guided generation for confirmed neurons.}
    \label{fig:Claim}
\end{figure}

\section{Effect of Image Degradation on Hidden-Layer Activations}\label{apd:degradation}

\begin{figure}[!t]
    \centering
    \includegraphics[width=16cm, height=7cm]{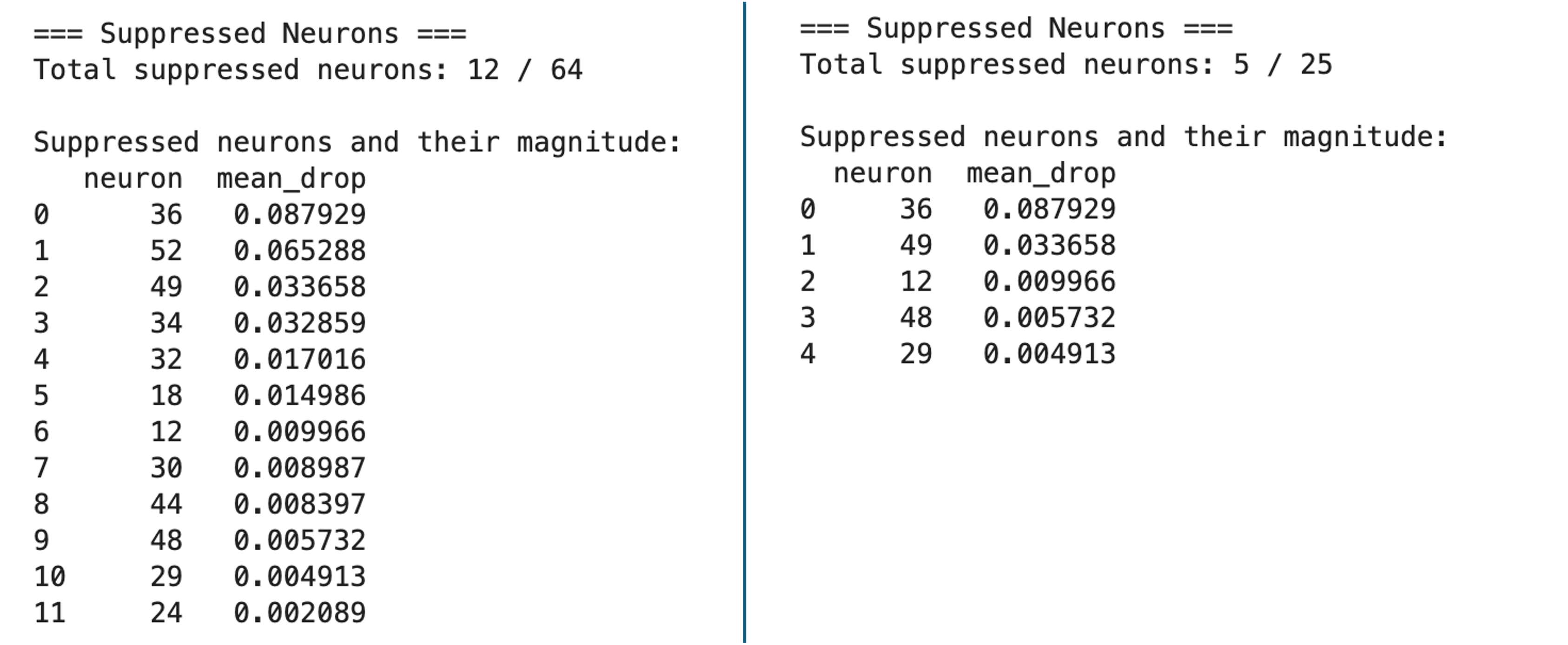}
    \caption{Suppression in neurons and their mean activation drop for JPEG compression. (a) Results for all neurons, (b) Results for confirmed neurons.}
    \label{fig:Claim_jpeg}
\end{figure}

As already noted in Section~\ref{rd}, the structure-guided generation setting can introduce visible distortions.
To test whether the observed activation differences are caused by image quality degradation, JPEG compression and Gaussian blur are applied to real images. Then the same hypothesis-testing procedure is repeated by comparing each real image with its transformed counterpart. 
For JPEG compression (q=70), the full real-fake activation suppression pattern is not reproduced. As observed from the results (depicted in \cref{hyp_1_jpeg_stats,hyp_2_jpeg_stat,hyp_3_jpeg_stats,hyp_4_jpeg_stats} and Figure~\ref{fig:Claim_jpeg}), $H_1$, annotation-based $H_2$, and $H_4$ are not supported. This means that JPEG-compressed real images did not show the same fake-like effect on the CNN hidden-layer activations. However, JPEG compression did affect some strongly activated neurons, as reflected in activation-based $H_2$ and $H_3$. This suggests that some activation sensitivity to degradation exists, but the effect is weaker and less semantically widespread than the fake-image effect.

\begin{table}[ht]
\footnotesize
\centering
\caption{Hypothesis 1 statistical results for real images with JPEG compression}
\label{hyp_1_jpeg_stats}

\begin{tabular}{
|>{\centering\arraybackslash}m{0.3\columnwidth}
|>{\centering\arraybackslash}m{0.45\columnwidth}|
}
\hline

\textbf{Statistic} &
\textbf{Real Images with JPEG Compression} \\
\hline

Images paired & 793 \\
\hline

Nonzero diffs used & 304 \\
\hline

% Mean firing count (real) & 0.803 \\
% \hline

% Mean firing count (fake) & 0.907 \\
% \hline

Median(real - compressed\_real) & -1.000 \\
\hline

Mean(real - compressed\_real) & -0.280 \\
\hline

Prop(real $>$ compressed\_real) & 0.418 \\
\hline

% Wilcoxon stat & $1.903550\times10^{4}$ \\
% \hline

$p$-value  & 0.9972669 \\
\hline

Effect size $r_{rb}$ & -0.164 \\
\hline

95\% bootstrap CI & [-0.49, -0.07] \\
 \hline
Decision & Fail to reject null hypothesis ($p$ $>$ 0.05) \\
\hline

\end{tabular}

\end{table}

\begin{table}[!htp]
\small
\centering
\caption{Hypothesis 2 statistical results for real images with JPEG compression}
\label{hyp_2_jpeg_stat}
\resizebox{\columnwidth}{!}{%
\begin{tabular}{
|>{\centering\arraybackslash}m{0.35\linewidth}
|>{\centering\arraybackslash}m{0.40\linewidth}
|>{\centering\arraybackslash}m{0.40\linewidth}|
}
\hline

\textbf{Statistic} &
\textbf{Concept Alignment} &
\textbf{Activation-Based Relevancy} \\
\hline

Images paired & 966 & 260 \\
\hline

Nonzero diffs used & 966 & 260 \\
\hline

Median(real - compressed\_real) & -0.0383 & 0.3315 \\
\hline

Mean(real - compressed\_real) & -0.0656 & 0.4086 \\
\hline

Prop(real - compressed\_real $>$ 0) & 0.478 & 0.677 \\
\hline

% Wilcoxon stat & 217517 & 25551 \\
% \hline

$p$-value & 0.9676 & $9.09 \times 10^{-13}$ \\
\hline

Effect size $r_{rb}$ & -0.069 & 0.504 \\
\hline

95\% bootstrap CI & [-0.12, -0.02] & [0.31, 0.51] \\
\hline

Decision & Fail to reject null hypothesis ($p$ $>$ 0.05) &
Reject null hypothesis ($p < 0.05$) \\
\hline

\end{tabular}
}
\end{table}

%\vspace{-1em}
\begin{table}[!htb]
\footnotesize
\centering
\caption{Hypothesis 3 statistical results for real images with JPEG compression}
\label{hyp_3_jpeg_stats}

\begin{tabular}{
|>{\centering\arraybackslash}m{0.45\columnwidth}
|>{\centering\arraybackslash}m{0.45\columnwidth}|
}
\hline

\textbf{Statistic} &
\textbf{Real Images with JPEG Compression} \\
\hline

Images paired & 793 \\
\hline

Nonzero diffs used & 168 \\
\hline

Mean(real\_count) & 0.803 \\
\hline

Mean(compressed\_real\_count) & 0.480 \\
\hline

Median(real\_count - compressed\_real\_count) & 1.000 \\
\hline

Prop(real\_count $>$ compressed\_real\_count) & 1.000 \\
\hline

% Wilcoxon stat & $1.402800 \times 10^{4}$ \\
% \hline

$p$-value & $5.943396 \times 10^{-32}$ \\
\hline

Effect size ($r_{rb}$) & 1.000 \\
\hline

95\% bootstrap CI & [1.39, 1.68] \\
\hline

Decision & Reject null hypothesis ($p < 0.05$) \\
\hline

\end{tabular}
\end{table}

\begin{table}[!htb]
\footnotesize
\centering
\caption{Hypothesis 4 statistical results for real images with JPEG compression}
\label{hyp_4_jpeg_stats}

\begin{tabular}{
|>{\centering\arraybackslash}m{0.45\columnwidth}
|>{\centering\arraybackslash}m{0.45\columnwidth}|
}
\hline

\textbf{Statistic} &
\textbf{Real Images with JPEG Compression} \\
\hline

Images paired & 793 \\
\hline

Nonzero diffs used & 151 \\
\hline

% Mean \#activated (real) & 0.328 \\
% \hline

% Mean \#activated (fake) & 0.354 \\
% \hline

Median(real - compressed\_real) counts & -1.000 \\
\hline

Prop(real $>$ compressed\_real) & 0.457 \\
\hline

% Wilcoxon stat & $5.173500 \times 10^{3}$ \\
% \hline

$p$-value & 0.8666343  \\
\hline

Effect size ($r_{rb}$) & -0.086 \\
\hline

95\% bootstrap CI & [-0.37, 0.1] \\
\hline

Decision & Fail to reject null hypothesis ($p > 0.05$) \\
\hline

\end{tabular}
\end{table}

\begin{figure}[!htb]
    \centering
    \includegraphics[width=16cm, height=11cm]{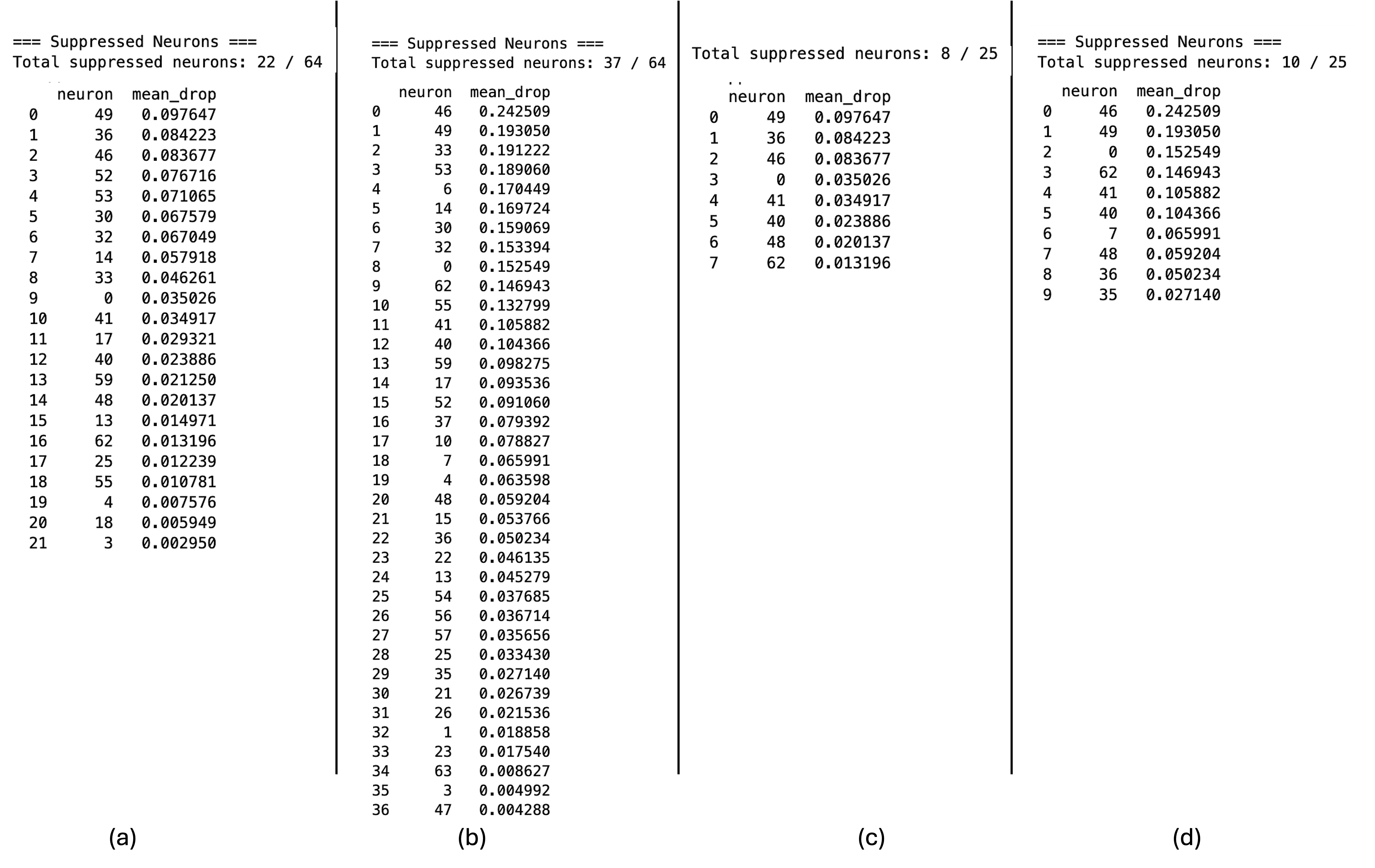}
    \caption{ Suppression in neurons and their mean activation drop. (a) Mild Blurred Image (all neurons), (b) Medium Blurred Image (all neurons), (c) Mild Blurred Image (confirmed neurons), and (d) Medium Blurred Image (confirmed neurons).}
    \label{fig:Claim_blur}
\end{figure}

%\vspace{-2em}
A similar pattern is observed for mild blur (kernel size, k=3 and sigma, s=1), results are depicted in \cref{hyp1_blur_stats,hyp2_blur_stats,hyp3_blur_stats,hyp4_blur_stats} and Figure~\ref{fig:Claim_blur}. The mildly blurred real images did not show fake-like behavior in the CNN hidden-layer activations, as they did not reproduce the full real-fake activation difference pattern. This further suggests that the hidden-layer activation differences observed for generated fake images cannot be explained by mild blurring alone.

%\vspace{-5em}
However, medium blur supported all hypotheses (results are illustrated in \cref{hyp1_blur_stats,hyp2_blur_stats,hyp3_blur_stats,hyp4_blur_stats} and Figure~\ref{fig:Claim_blur}), indicating that stronger degradation can induce fake-like hidden-layer behavior. This reveals a limitation of the proposed approach: when real images undergo substantial visual degradation, their hidden-layer activation patterns may become similar to those observed for generated fake images. This is expected because medium blur removes edge, texture, and local detail, and CNN activations are sensitive to such visual information loss.
However, the p-values for medium blur are larger than those observed for generated fake images, suggesting that the fake image effect is stronger in fake image analysis experiments.

Therefore, the proposed method is not fully invariant to strong image quality degradation. Overall, the results suggest that mild blur and medium JPEG compression (q=70) degradation do not fully produce effects similar to those observed in the real-fake comparison, while stronger degradation such as medium blur can potentially confound the analysis.
%\vspace{-1em}
\begin{table}[h]
\small
\centering
\caption{Hypothesis 1 statistical results for real images with mild and medium blurred real images}
\label{hyp1_blur_stats}
\resizebox{\columnwidth}{!}{%
\begin{tabular}{
|>{\centering\arraybackslash}m{0.30\linewidth}
|>{\centering\arraybackslash}m{0.40\linewidth}
|>{\centering\arraybackslash}m{0.40\linewidth}|
}
\hline

\textbf{Statistic} &
\textbf{Mild Blur} &
\textbf{Medium Blur} \\
\hline

Images paired & 793 & 793 \\
\hline

Nonzero diffs used & 295 & 302 \\
\hline

% Mean firing count (real) & 0.803 & 0.803 \\
% \hline

% Mean firing count (fake) & 0.820 & 0.677 \\
% \hline

Median(real - blurred\_real) & 1.000 & 1.000 \\
\hline

Mean(real - blurred\_real) & -0.044 & 0.331 \\
\hline

Prop(real $>$ blurred\_real) & 0.515 & 0.606 \\
\hline

% Wilcoxon stat & $2.143750 \times 10^{4}$ & $2.772700 \times 10^{4}$ \\
% \hline

$p$-value & 0.6090308 & $4.809483 \times 10^{-4}$ \\
\hline

Effect size ($r_{rb}$) & 0.031 & 0.212 \\
\hline

95\% bootstrap CI & [-0.26, 0.17] & [0.12, 0.54] \\
\hline

Decision & Fail to reject null hypothesis ($p > 0.05$) &
Reject null hypothesis ($p < 0.05$) \\
\hline

\end{tabular}
}
\end{table}

\begin{table}[!htb]

\centering
\caption{Hypothesis 2 statistical results for real images with mild and medium blurred real images}
\label{hyp2_blur_stats}

\resizebox{\columnwidth}{!}{%
\begin{tabular}{
|>{\centering\arraybackslash}m{0.32\linewidth}
|>{\centering\arraybackslash}m{0.25\linewidth}
|>{\centering\arraybackslash}m{0.25\linewidth}
|>{\centering\arraybackslash}m{0.25\linewidth}
|>{\centering\arraybackslash}m{0.25\linewidth}|
}
\hline

\multirow{2}{*}{\textbf{Statistic}} &
\multicolumn{2}{c|}{\textbf{Mild Blur}} &
\multicolumn{2}{c|}{\textbf{Medium Blur}} \\
\cline{2-5}

&
\textbf{Concept Alignment } &
\textbf{Activation-Based Relevancy} &
\textbf{Concept Alignment } &
\textbf{Activation-Based Relevancy} \\
\hline

N total & 967 & 260 & 960 & 260 \\
\hline

Nonzero diffs used & 967 & 260 & 960 & 260 \\
\hline

Median(real - blurred\_real) & 0.0226 & 0.4866 & 0.1731 & 0.7375 \\
\hline

Mean(real - blurred\_real) & 0.0261 & 0.5052 & 0.1861 & 0.7686 \\
\hline

Prop(real $>$ blurred\_real) & 0.508 & 0.754 & 0.571 & 0.762 \\
\hline

% Wilcoxon stat & 241963 & 27160 & 277974 & 28683 \\
% \hline

$p$-value & 0.1801  & $2.237 \times 10^{-17}$ & $1.812 \times 10^{-8}$ & $2.350 \times 10^{-22}$ \\
\hline

Effect size $r_{rb}$ & 0.034 & 0.601 & 0.205 & 0.691 \\
\hline

95\% bootstrap CI &  [-0.03, 0.08] & [0.4, 0.61] & [0.12, 0.25] & [0.64, 0.9] \\
\hline

Decision & Fail to reject null hypothesis ($p > 0.05$) &
Reject null hypothesis ($p < 0.05$) &
Reject null hypothesis ($p < 0.05$) &
Reject null hypothesis ($p < 0.05$) \\
\hline

\end{tabular}
}
\end{table}

\begin{table}[!htb]
\small
\centering
\caption{Hypothesis 3 statistical results for real images with mild and medium blurred real
images}
\label{hyp3_blur_stats}

\resizebox{\columnwidth}{!}{%
\begin{tabular}{
|>{\centering\arraybackslash}m{0.4\linewidth}
|>{\centering\arraybackslash}m{0.35\linewidth}
|>{\centering\arraybackslash}m{0.35\linewidth}|
}
\hline

\textbf{Statistic} &
\textbf{Mild Blur} &
\textbf{Medium Blur} \\
\hline

Images paired & 793 & 793 \\
\hline

Nonzero diffs used & 189 & 229 \\
\hline

% Mean(real\_count) & 0.803 & 0.803 \\
% \hline

% Mean(fake\_count) & 0.445 & 0.348 \\
% \hline

Median(real\_count - blurred\_real\_count) & 1.000 & 1.000 \\
\hline

Prop(real\_count $>$ blurred\_real\_count) & 1.000 & 1.000 \\
\hline

% Wilcoxon stat & $1.795500 \times 10^{4}$ & $2.633500 \times 10^{4}$ \\
% \hline

$p$-value & $1.085851 \times 10^{-35}$ & $2.100855 \times 10^{-42}$ \\
\hline

Effect size ($r_{rb}$) & 1.000 & 1.000 \\
\hline

95\% bootstrap CI &  [1.38, 1.63] & [1.45, 1.72] \\
\hline

Decision & Reject null hypothesis ($p < 0.05$) &
Reject null hypothesis ($p < 0.05$) \\
\hline

\end{tabular}
}
\end{table}

\begin{table}[!htb]
\small
\centering
\caption{Hypothesis 4 statistical results for real images with mild and medium blurred real
images}
\label{hyp4_blur_stats}

\resizebox{\columnwidth}{!}{%
\begin{tabular}{
|>{\centering\arraybackslash}m{0.3\linewidth}
|>{\centering\arraybackslash}m{0.4\linewidth}
|>{\centering\arraybackslash}m{0.35\linewidth}|
}
\hline

\textbf{Statistic} &
\textbf{Mild Blur} &
\textbf{Medium Blur} \\
\hline

Images paired & 793 & 793 \\
\hline

Nonzero diffs used & 155 & 162 \\
\hline

% Mean \#activated (real) & 0.328 & 0.328 \\
% \hline

% Mean \#activated (fake) & 0.325 & 0.266 \\
% \hline

Median(real - blurred\_real) & 1.000 & 1.000 \\
\hline

Prop(real $>$ blurred\_real) & 0.523 & 0.605 \\
\hline

% Wilcoxon stat & $6.192500 \times 10^{3}$ & $8.240000 \times 10^{3}$ \\
% \hline

$p$-value & 0.3883456  & $1.639224 \times 10^{-3}$ \\
\hline

Effect size ($r_{rb}$) & 0.045 & 0.210 \\
\hline

95\% bootstrap CI & [-0.21, 0.23] & [0.08, 0.52]\\
\hline

Decision & Fail to reject null hypothesis ($p > 0.05$) &
Reject null hypothesis ($p < 0.05$) \\
\hline

\end{tabular}
}
\end{table}

% This is the first appendix.

% \newpage
% \clearpage
%\section{Second Appendix}\label{apd:second}
% \vspace{5em}
\section{Generalizability Analysis Using Alternative Dataset, CNN Architectures, and Image Generators}\label{apd:second}

To examine whether the observed hidden-neuron activation differences are limited to the initial experiment with InceptionV3, SUN2012, and Stable Diffusion, additional validation experiments are conducted under an alternative experimental configuration. Specifically, ResNet50V2 is used as the CNN architecture, ADE20K as the real-image dataset, and FLUX 0.1 as an additional image generation model. The choice of ResNet50V2 and ADE20K is motivated by prior work~\citep{10.1007/978-3-031-71170-1_12} on the Concept Induction-Based Neuron Interpretability Framework. Therefore, this setting allows to evaluate the proposed real-fake activation analysis in a configuration that is both different from the initial experiment and grounded in an existing neuron-interpretability setup. 

\begin{figure}[!htb]
    \centering
    \includegraphics[width=16cm, height=10cm]{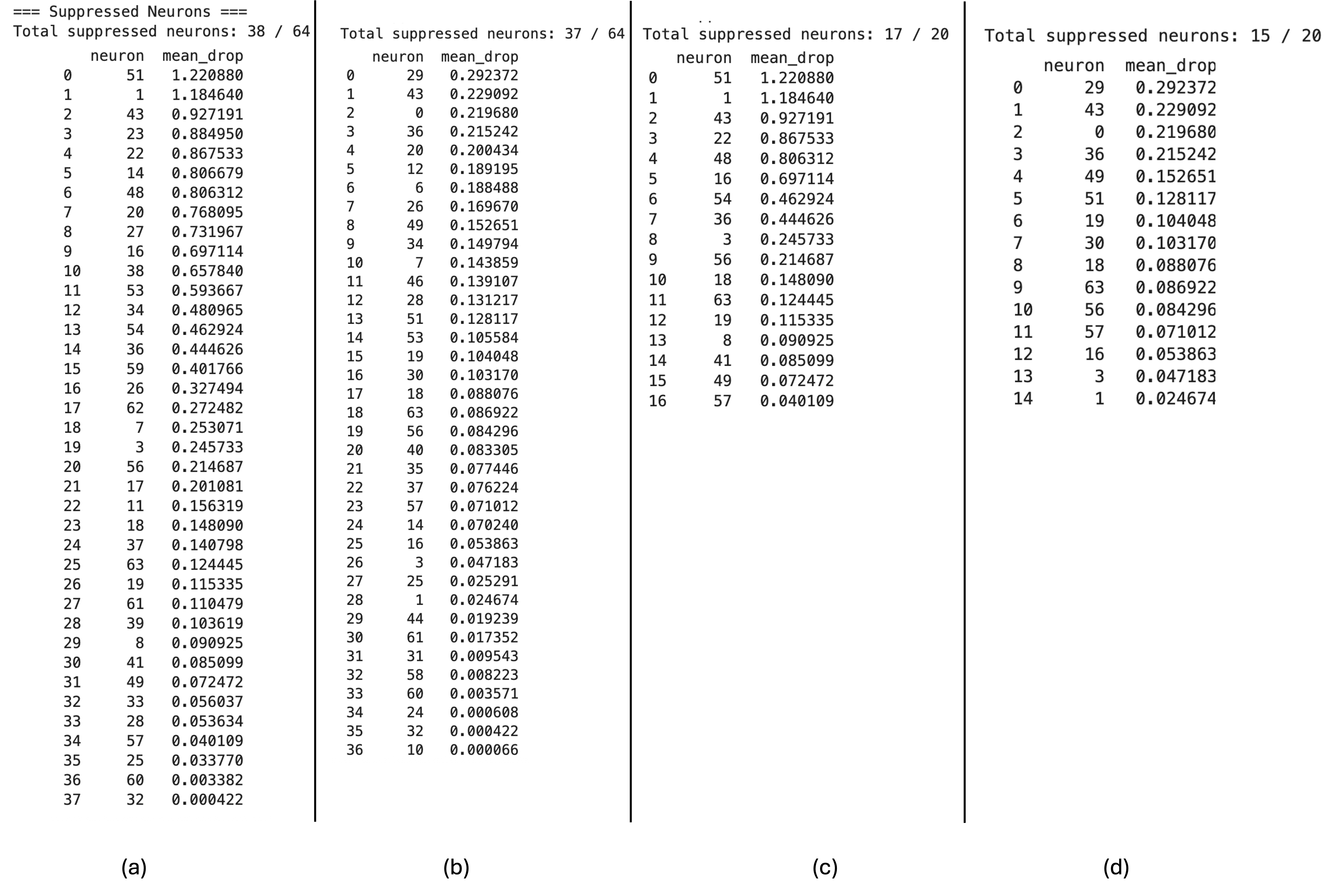}
    \caption{Suppression in neurons and their mean activation drop with ADE20K dataset, ResNet50V2 CNN model, and Stable Diffusion image generation model. (a) Text-guided fake image generation (all neurons), (b) Structure-guided fake image generation (all neurons), (c) Text-guided generation for confirmed neurons, and (d) Structure-guided generation for confirmed neurons.}
    \label{fig:Claim_ade_stab}
\end{figure}

\begin{figure}[H]
    \centering
    \includegraphics[height=11cm,width=15cm]{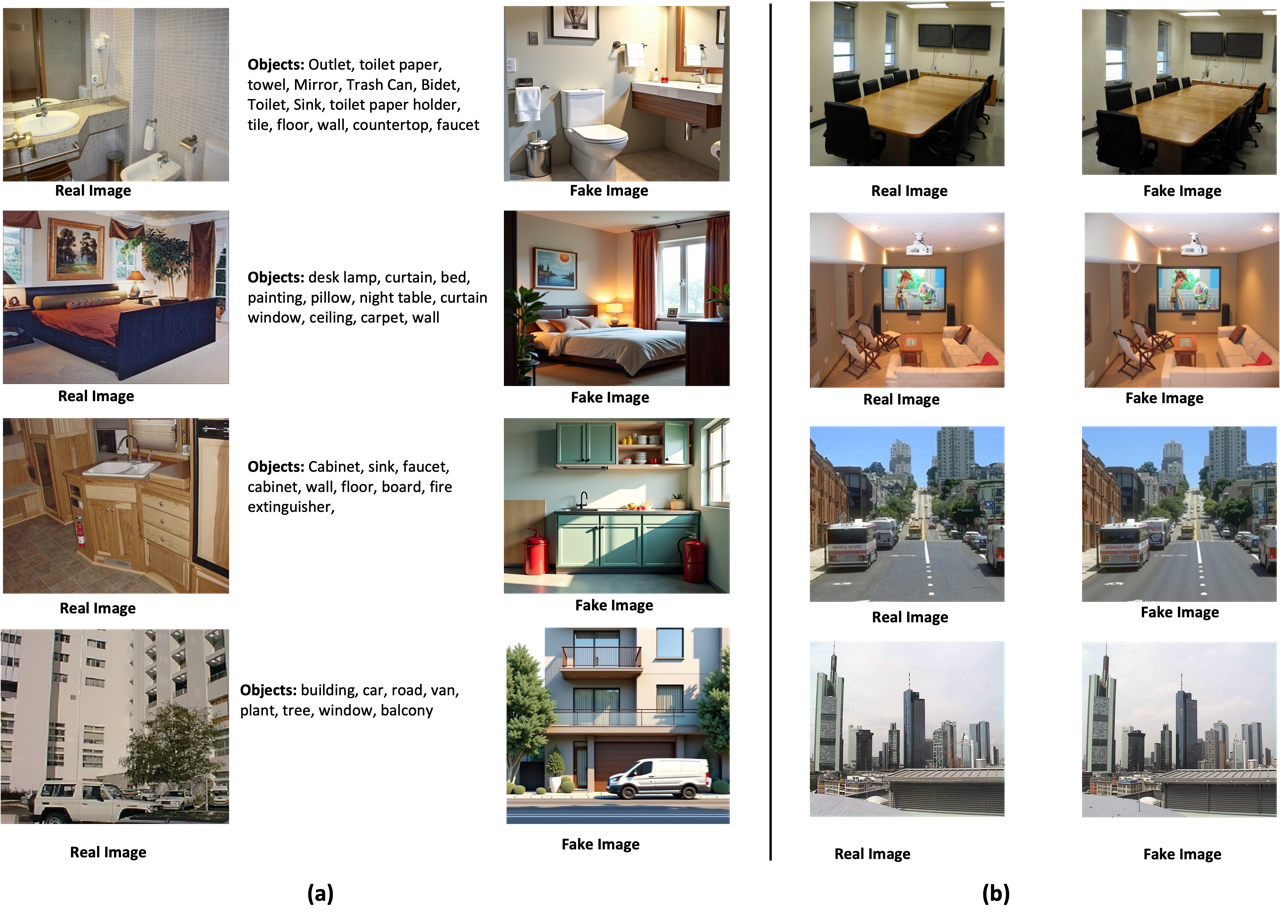}
    \caption{Additional Sample fake output images using ADE20K and Stable Diffusion variants; (a): from Text-Guided Fake Image Generation; (b): from Structure-Guided Fake Image Generation.}
    \label{fig:fake_sample_ade_stab}
\end{figure}

For ADE20K with ResNet50V2 as the CNN model and Stable Diffusion as the image generation model, all hypotheses are supported in both generation settings, namely text-guided generation and structure-guided generation. Significant neuron activation suppression is also observed for the claim in both settings, as shown in \cref{hyp1_ade_stab,hyp2_ade_stab,hyp3_ade_stab,hyp4_ade_stab} and Figure~\ref{fig:Claim_ade_stab}. Sample of the generated fake images of this setup is depicted in Figure~\ref{fig:fake_sample_ade_stab}. 

\begin{figure}[!t]
    \centering
    \includegraphics[width=16cm, height=10cm]{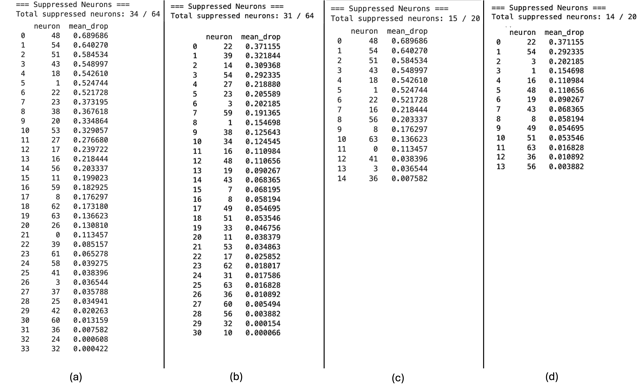}
    \caption{Suppression in neurons and their mean activation drop with ADE20K dataset, ResNet50V2 CNN model, and FLUX image generation model. (a) Text-guided fake image generation (all neurons), (b) Structure-guided fake image generation (all neurons), (c) Text-guided generation for confirmed neurons, and (d) Structure-guided generation for confirmed neurons.}
    \label{fig:Claim_ade_flux}
\end{figure}

\begin{figure}[H]
    \centering
    \includegraphics[width=\textwidth]{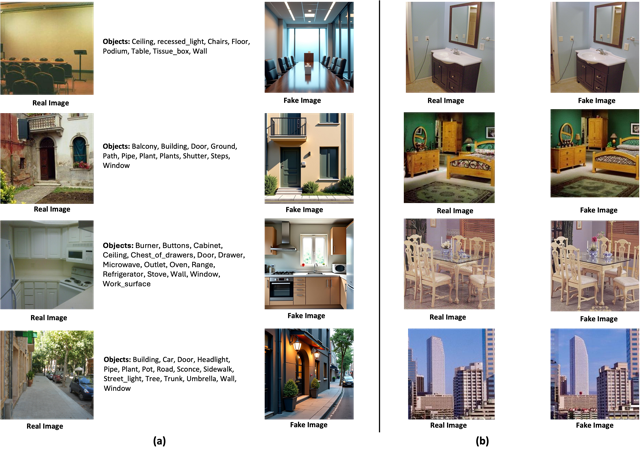}
    \caption{Additional Sample fake output images using ADE20K and FLUX; (a): from Text-Guided Fake Image Generation; (b): from Structure-Guided Fake Image Generation.}
    \label{fig:fake_sample_ade_flux_}
\end{figure}

Moreover, ADE20K with ResNet50V2 and FLUX is also evaluated under the same two generation settings. As shown in \cref{hyp1_ade_flux,hyp2_ade_flux,hyp3_ade_flux,hyp4_ade_flux}, all four hypotheses are supported for both text-guided and structure-guided fake image generation. Significant neuron activation suppression is also observed for both all-neuron and confirmed-neuron groups, as shown in Figure~\ref{fig:Claim_ade_flux}. Sample of the generated fake images of this setup is depicted in Figure~\ref{fig:fake_sample_ade_flux_}.

These results suggest that the proposed hidden-neuron activation differences are not limited to the initially evaluated SUN2012, InceptionV3, and Stable Diffusion experiment, but can also be observed with a different dataset, CNN architecture, and image generation model.

\begin{table}[!htb]
\small
\centering
\caption{Hypothesis 1 statistical results with ADE20K dataset, ResNet50V2 CNN model, and Stable Diffusion image generation model.}
\label{hyp1_ade_stab}

\resizebox{\columnwidth}{!}{%
\begin{tabular}{
|>{\centering\arraybackslash}m{0.24\linewidth}
|>{\centering\arraybackslash}m{0.45\linewidth}
|>{\centering\arraybackslash}m{0.5\linewidth}|
}
\hline

\textbf{Statistic} &
\textbf{Text-Guided Fake Image Generation} &
\textbf{Structure-Guided Fake Image Generation} \\
\hline

Images paired & 1370 & 1370 \\
\hline

Nonzero diffs used & 399 & 339 \\
\hline

% Mean firing count (real) & 0.380 & 0.422 \\
% \hline

% Mean firing count (fake) & 0.212 & 0.370 \\
% \hline

Median(real-fake) & 1.000 & 1.000 \\
\hline

Mean(real-fake) & 0.574 & 0.209 \\
\hline

Prop(real $>$ fake) & 0.667 & 0.558 \\
\hline

% Wilcoxon statistic & $5.397350\times10^{4}$ & $3.269700\times10^{4}$ \\
% \hline

$p$-value & $1.452988\times10^{-10}$ & $1.203812\times10^{-2}$ \\
\hline

Effect size $r_{rb}$ & 0.333 & 0.115 \\
\hline

95\% bootstrap CI & [0.38, 0.77] & [0.03, 0.39]\\
\hline

Decision & Reject null hypothesis ($p < 0.05$) &
Reject null hypothesis ($p < 0.05$) \\
\hline

\end{tabular}
}
\end{table}

\begin{table}[!htb]
\small
\centering
\caption{Hypothesis 2 statistical results with ADE20K dataset, ResNet50V2 CNN model, and Stable Diffusion image generation model.}
\label{hyp2_ade_stab}

\resizebox{\columnwidth}{!}{%
\begin{tabular}{
|>{\centering\arraybackslash}m{0.20\linewidth}
|>{\centering\arraybackslash}m{0.25\linewidth}
|>{\centering\arraybackslash}m{0.25\linewidth}
|>{\centering\arraybackslash}m{0.25\linewidth}
|>{\centering\arraybackslash}m{0.30\linewidth}|
}
\hline

\textbf{Statistic} &
\textbf{Text-Guided Fake Generation (Concept Alignment)} &
\textbf{Structure-Guided Fake Generation (Concept Alignment)} &
\textbf{Text-Guided Fake Generation (Activation-Based Relevancy)} &
\textbf{Structure-Guided Fake Generation (Activation-Based
Relevancy)} \\
\hline

Images paired & 888 & 880 & 269 & 285 \\
\hline

Nonzero diffs used & 888 & 880 & 269 & 285 \\
\hline

Median(real - fake) & 1.4678 & 0.1044 & 4.2830 & 0.8398 \\
\hline

Mean(real - fake) & 1.3775 & 0.4081 & 4.3264 & 1.4776 \\
\hline

Prop(real $>$ fake) & 0.718 & 0.536 & 0.970 & 0.775 \\
\hline

% Wilcoxon statistic & 309519 & 229862 & 36203 & 35311 \\
% \hline

$p$-value & $4.977\times10^{-49}$ & $8.824\times10^{-7}$ & $1.252\times10^{-45}$ & $3.940\times10^{-27}$ \\
\hline

Effect size ($r_{rb}$) & 0.568 & 0.186 & 0.994 & 0.733 \\
\hline

95\% bootstrap CI & [1.12, 1.43] & [0.34, 0.54]  & [3.92, 4.45] & [1.15, 1.61]\\
\hline

Decision & Reject null hypothesis ($p < 0.05$) &
Reject null hypothesis ($p < 0.05$) &
Reject null hypothesis ($p < 0.05$) &
Reject null hypothesis ($p < 0.05$) \\
\hline

\end{tabular}
}
\end{table}

\begin{table}[hbt]
\small
\centering
\caption{Hypothesis 3 statistical results with ADE20K dataset, ResNet50V2 CNN model, and Stable Diffusion image generation model.}
\label{hyp3_ade_stab}

\resizebox{\columnwidth}{!}{%
\begin{tabular}{
|>{\centering\arraybackslash}m{0.32\linewidth}
|>{\centering\arraybackslash}m{0.42\linewidth}
|>{\centering\arraybackslash}m{0.48\linewidth}|
}
\hline

\textbf{Statistic} &
\textbf{Text-Guided Fake Image Generation} &
\textbf{Structure-Guided Fake Image Generation} \\
\hline

Images paired & 1370 & 1370 \\
\hline

Nonzero diffs used & 329 & 215 \\
\hline

Mean (real\_count) & 0.431 & 0.431 \\
\hline

Mean (fake\_count) & 0.014 & 0.196 \\
\hline

Median(real\_count - fake\_count) & 1.000 & 1.000 \\
\hline

Prop(real\_count $>$ fake\_count) & 1.000 & 1.000 \\
\hline

% Wilcoxon statistic & $5.428500\times10^{4}$ & $2.322000\times10^{4}$ \\
% \hline

$p$-value & $4.144095\times10^{-59}$ & $1.433383\times10^{-40}$ \\
\hline

Effect size ($r_{rb}$) & 1.000 & 1.000 \\
\hline

95\% bootstrap CI & [1.62, 1.87] &  [1.39, 1.63]\\
\hline

Decision & Reject null hypothesis ($p < 0.05$) &
Reject null hypothesis ($p < 0.05$) \\
\hline

\end{tabular}
}
\end{table}

\begin{table}[hbt]
\small
\centering
\caption{Hypothesis 4 statistical results with ADE20K dataset, ResNet50V2 CNN model, and Stable Diffusion image generation model.}
\label{hyp4_ade_stab}

\resizebox{\columnwidth}{!}{%
\begin{tabular}{
|>{\centering\arraybackslash}m{0.35\linewidth}
|>{\centering\arraybackslash}m{0.45\linewidth}
|>{\centering\arraybackslash}m{0.50\linewidth}|
}
\hline

\textbf{Statistic} &
\textbf{Text-Guided Fake Image Generation} &
\textbf{Structure-Guided Fake Image Generation} \\
\hline

Images paired & 1370 & 1370 \\
\hline

Nonzero diffs used & 232 & 205 \\
\hline

% Mean \#activated (real) & 0.216 & 0.230 \\
% \hline

% Mean \#activated (fake) & 0.064 & 0.189 \\
% \hline

Median (real-fake) counts & 1.000 & 1.000 \\
\hline

Prop(real $>$ fake) & 0.828 & 0.546 \\
\hline

% Wilcoxon statistic & $2.229000\times10^{4}$ & $1.236550\times10^{4}$ \\
% \hline

$p$-value & $1.044829\times10^{-19}$ & $1.176608\times10^{-2}$ \\
\hline

Effect size ($r_{rb}$) & 0.655 & 0.093 \\
\hline

95\% bootstrap CI & [0.72, 1.08] &  [0.07, 0.47]\\
\hline

Decision & Reject null hypothesis ($p < 0.05$) &
Reject null hypothesis ($p < 0.05$) \\
\hline

\end{tabular}
}
\end{table}

% In case of structure-guided generation of Flux with ADE20k and ResNet50v2, the results are more nuanced, as observed in \cref{hyp1_ade_flux,hyp2_ade_flux,hyp3_ade_flux,hyp4_ade_flux}. Here, both Hypothesis 3 and activation-based relevancy in Hypothesis 2 are supported. However, Hypothesis 1 and Hypothesis 4 are not supported, and annotation-based relevance in Hypothesis 2 is also not statistically significant. In these cases, the effect-size trends are small: Hypothesis 1 and annotation-based Hypothesis 2 showed weak positive directions favoring real images, while Hypothesis 4 showed a very small negative direction favoring fake images. Moreover, as observed from Figure~\ref{fig:Claim_ade_flux}, Flux-generated images in this setup still produce activation suppression in both all-neuron and confirmed-neuron groups, with stronger suppression in text-guided generation than in structure-guided generation.

\begin{table}[!htb]
\small
\centering
\caption{Hypothesis 1 statistical results with ADE20K dataset, ResNet50V2 CNN model, and FLUX image generation model.}
\label{hyp1_ade_flux}

\resizebox{\columnwidth}{!}{%
\begin{tabular}{
|>{\centering\arraybackslash}m{0.25\linewidth}
|>{\centering\arraybackslash}m{0.45\linewidth}
|>{\centering\arraybackslash}m{0.50\linewidth}|
}
\hline

\textbf{Statistic} &
\textbf{Text-Guided Fake Image Generation} &
\textbf{Structure-Guided Fake Image Generation} \\
\hline

Images paired & 1370 & 1370 \\
\hline

Nonzero diffs used & 400 & 329 \\
\hline

% Mean firing count (real) & 0.413 & 0.429 \\
% \hline

% Mean firing count (fake) & 0.174 & 0.299 \\
% \hline

Median(real - fake) & 1.000 & 1.000 \\
\hline

Mean(real - fake) & 0.818 & 0.356 \\
\hline

Prop(real $>$ fake) & 0.718 & 0.626 \\
\hline

% Wilcoxon statistic & $5.895900\times10^{4}$ & $4.030300\times10^{4}$ \\
% \hline

$p$-value & $2.304215\times10^{-17}$ & $3.436013\times10^{-5} $\\
\hline

Effect size ($r_{rb}$) & 0.435 & 0.252 \\
\hline

95\% bootstrap CI & [0.64, 1.01] &  [0.17, 0.53]\\
\hline

Decision & Reject null hypothesis ($p < 0.05$) &
Reject null hypothesis ($p < 0.05$) \\
\hline

\end{tabular}
}
\end{table}

%\vspace{-10em}
\begin{table}[!htb]
\small
\centering
\caption{Hypothesis 2 statistical results with ADE20K dataset, ResNet50V2 CNN model, and FLUX image generation model.}
\label{hyp2_ade_flux}

\resizebox{\columnwidth}{!}{%
\begin{tabular}{
|>{\centering\arraybackslash}m{0.20\linewidth}
|>{\centering\arraybackslash}m{0.25\linewidth}
|>{\centering\arraybackslash}m{0.3\linewidth}
|>{\centering\arraybackslash}m{0.25\linewidth}
|>{\centering\arraybackslash}m{0.30\linewidth}|
}
\hline

\textbf{Statistic} &
\textbf{Text-Guided Fake Image Generation (Concept Alignment)} &
\textbf{Structure-Guided Fake Image Generation (Concept Alignment)} &
\textbf{Text-Guided Fake Image Generation (Activation-Based Relevancy)} &
\textbf{Structure-Guided Fake Image Generation (Activation-Based Relevancy)} \\
\hline

Images paired & 925 & 905 & 316 & 292 \\
\hline

Nonzero diffs used & 925 & 905 & 316 & 292 \\
\hline

Median(real - fake) & 1.1181 & 0.1640 & 3.3038 & 0.8045 \\
\hline

Mean(real - fake) & 1.0456 & 0.1993 & 3.4708 & 0.8583 \\
\hline

Prop(real $>$ fake) & 0.692 & 0.572 & 0.984 & 0.757 \\
\hline

% Wilcoxon statistic & 326032 & 208131 & 50030 & 32355 \\
% \hline

$p$-value & $2.017\times10^{-43}$ & $1.966\times10^{-8}$ & $1.251\times10^{-53}$ & $8.404\times10^{-24}$ \\
\hline

Effect size ($r_{rb}$) & 0.523 & 0.211 & 0.998 & 0.675 \\
\hline

95\% bootstrap CI & [0.91, 1.18] & [0.13, 0.27] & [3.27, 3.68] & [0.72, 1.0] \\
\hline

Decision & Reject null hypothesis ($p < 0.05$) &
Reject null hypothesis ($p < 0.05$) &
Reject null hypothesis ($p < 0.05$) &
Reject null hypothesis ($p < 0.05$) \\
\hline
\end{tabular}
}
\end{table}

%\vspace{-5em}
\begin{table}[!hbt]
\small
\centering
\caption{Hypothesis 3 statistical results with ADE20K dataset, ResNet50V2 CNN model, and FLUX image generation model.}
\label{hyp3_ade_flux}

\resizebox{\columnwidth}{!}{%
\begin{tabular}{
|>{\centering\arraybackslash}m{0.35\linewidth}
|>{\centering\arraybackslash}m{0.45\linewidth}
|>{\centering\arraybackslash}m{0.50\linewidth}|
}
\hline

\textbf{Statistic} &
\textbf{Text-Guided Fake Image Generation} &
\textbf{Structure-Guided Fake Image Generation} \\
\hline

Images paired & 1370 & 1370 \\
\hline

Nonzero diffs used & 321 & 229 \\
\hline

Mean (real\_count) & 0.431 & 0.431 \\
\hline

Mean (fake\_count) & 0.021 & 0.191 \\
\hline

Median(real\_count - fake\_count) & 1.000 & 1.000 \\
\hline

Prop(real\_count $>$ fake\_count) & 1.000 & 1.000 \\
\hline

% Wilcoxon statistic & $5.168100\times10^{4}$ & $2.030100\times10^{4}$ \\
% \hline

$p$-value & $1.368160\times10^{-57}$ & $3.32\times10^{-43}$ \\
\hline

Effect size ($r_{rb}$) & 1.000 & 1.000 \\
\hline

95\% bootstrap CI & [1.63, 1.88] &  [1.34, 1.55]\\
\hline

Decision & Reject null hypothesis ($p < 0.05$) &
Reject null hypothesis ($p < 0.05$) \\
\hline

\end{tabular}
}
\end{table}

\begin{table}[!htb]
\small
\centering
\caption{Hypothesis 4 statistical results with ADE20K dataset, ResNet50V2 CNN model, and FLUX image generation model.}
\label{hyp4_ade_flux}

\resizebox{\columnwidth}{!}{%
\begin{tabular}{
|>{\centering\arraybackslash}m{0.35\linewidth}
|>{\centering\arraybackslash}m{0.45\linewidth}
|>{\centering\arraybackslash}m{0.50\linewidth}|
}
\hline

\textbf{Statistic} &
\textbf{Text-Guided Fake Image Generation} &
\textbf{Structure-Guided Fake Image Generation} \\
\hline

Images paired & 1370 & 1370 \\
\hline

Nonzero diffs used & 256 & 212 \\
\hline

% Mean \#activated (real) & 0.231 & 0.235 \\
% \hline

% Mean \#activated (fake) & 0.085 & 0.239 \\
% \hline

Median(real - fake) & 1.000 & 1.000 \\
\hline

Prop(real $>$ fake) & 0.758 & 0.580 \\
\hline

% Wilcoxon statistic & $2.504400\times10^{4}$ & $1.067250\times10^{4}$ \\
% \hline

$p$-value & $1.963471\times10^{-14}$ & $2.452926\times10^{-2}$ \\
\hline

Effect size ($r_{rb}$) & 0.516 & 0.160 \\
\hline

95\% bootstrap CI & [0.59, 0.97] &  [-0.01, 0.38]\\
\hline

Decision & Reject null hypothesis ($p < 0.05$) &
Reject null hypothesis ($p < 0.05$) \\
\hline

\end{tabular}
}
\end{table}

\clearpage
% \section{Third Appendix}\label{apd:third}
\subsection{Generated Image Samples Across Models}
\label{apd:samples}
\begin{figure}[H]
    \centering
    \includegraphics[height=11cm,width=15cm]{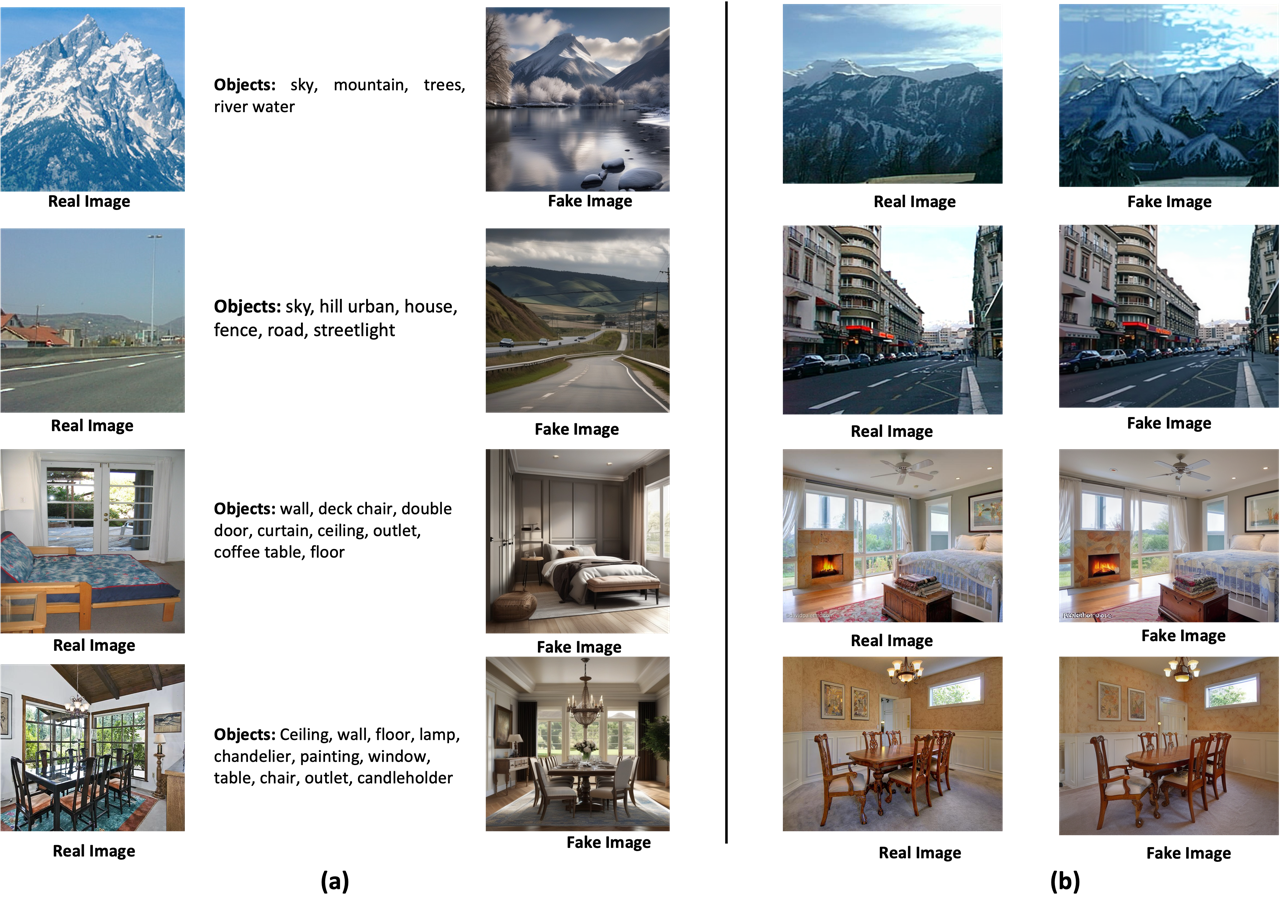}
    \caption{Additional Sample fake output images using SUN2012 and Stable Diffusion variants; (a): from Text-Guided Fake Image Generation; (b): from Structure-Guided Fake Image Generation.}
    \label{fig:fake_sample_sun_stab}
\end{figure}

% \clearpage
\end{document}